%% file: main.tex
\documentclass[runningheads]{llncs}
\usepackage{graphicx}

\usepackage{tikz}
\usepackage{comment}
\usepackage{amsmath,amssymb} 
\usepackage{color}
\usepackage{graphicx}
\usepackage[hidelinks]{hyperref}
\usepackage{amsmath,amssymb} 
\usepackage{color}
\usepackage[font=small,labelfont=bf,tableposition=top]{caption}
\usepackage[font=footnotesize]{subcaption}
\usepackage{booktabs}
\usepackage{tablefootnote}
\usepackage{threeparttable}
\usepackage{multirow}
\usepackage{arydshln}
\usepackage[accsupp]{axessibility}  

\begin{document}
\pagestyle{headings}
\mainmatter

\title{Generalized Few-Shot Semantic Segmentation: All You Need is Fine-Tuning}

\author{Josh Myers-Dean\textsuperscript{*}
Yinan Zhao\textsuperscript{+}
Brian Price\textsuperscript{\textdagger}
Scott Cohen\textsuperscript{\textdagger}
Danna Gurari\textsuperscript{*,+}}
\institute{\small {}\textsuperscript{*}University of Colorado, Boulder\quad {}\textsuperscript{+}University of Texas at Austin\quad {}\textsuperscript{\textdagger}Adobe Research
}

\maketitle

\input{abstract}
\vspace{-.5em}
\input{introduction}
\vspace{-.5em}
\input{related_work}
\vspace{-.5em}
\input{method}
\vspace{-.5em}
\input{experiments}

\vspace{-.5em}
\input{conclusion}
\vspace{-.5em}
\paragraph{Acknowledgments.}
We gratefully acknowledge support from Microsoft AI for Accessibility for donating cloud computing credits and Samreen Anjum for formatting help. This work is also supported by the National Science Foundation Graduate Research Fellowship Program under Grant No. DGE 2040434.  
\clearpage

%
%
\newpage
\bibliographystyle{splncs04}
\bibliography{egbib}
\clearpage
\input{supplementary}
\end{document}

%% file: abstract.tex
\begin{abstract}
Generalized few-shot semantic segmentation was introduced to move beyond only evaluating few-shot segmentation models on novel classes to include testing their ability to remember base classes. While the current state-of-the-art approach is based on meta-learning, it performs poorly and saturates in learning after observing only a few shots. We propose the first fine-tuning solution, and demonstrate that it addresses the saturation problem while achieving state-of-the-art results on two datasets, PASCAL-$5^i$ and COCO-$20^i$. We also show that it outperforms existing methods, whether fine-tuning multiple final layers or only the final layer. Finally, we present a triplet loss regularization that shows how to redistribute the balance of performance between novel and base categories so that there is a smaller gap between them.
\end{abstract}

%% file: introduction.tex
\section{Introduction}
Few-shot semantic segmentation is the task of learning to segment select categories in an image when given only a limited number of annotated examples for each category.  Generally, such methods first train on base classes for which there is an abundance of labeled data and then try to generalize to novel classes for which only few annotated examples are available. A limitation of most models is that they are only evaluated with respect to their performance on the novel classes~\cite{wang2019panet,zhang2019canet,boudiaf2021fewshot}. In other words, existing work largely ignores whether the model retains knowledge of how to analyze the base categories.  To combat this limitation, generalized few-shot semantic segmentation was introduced as an extension of few-shot semantic segmentation in 2020, such that the performance of models are evaluated on both the base classes and novel classes.

\begin{figure}[!t]
\centering
\includegraphics[width=\linewidth]{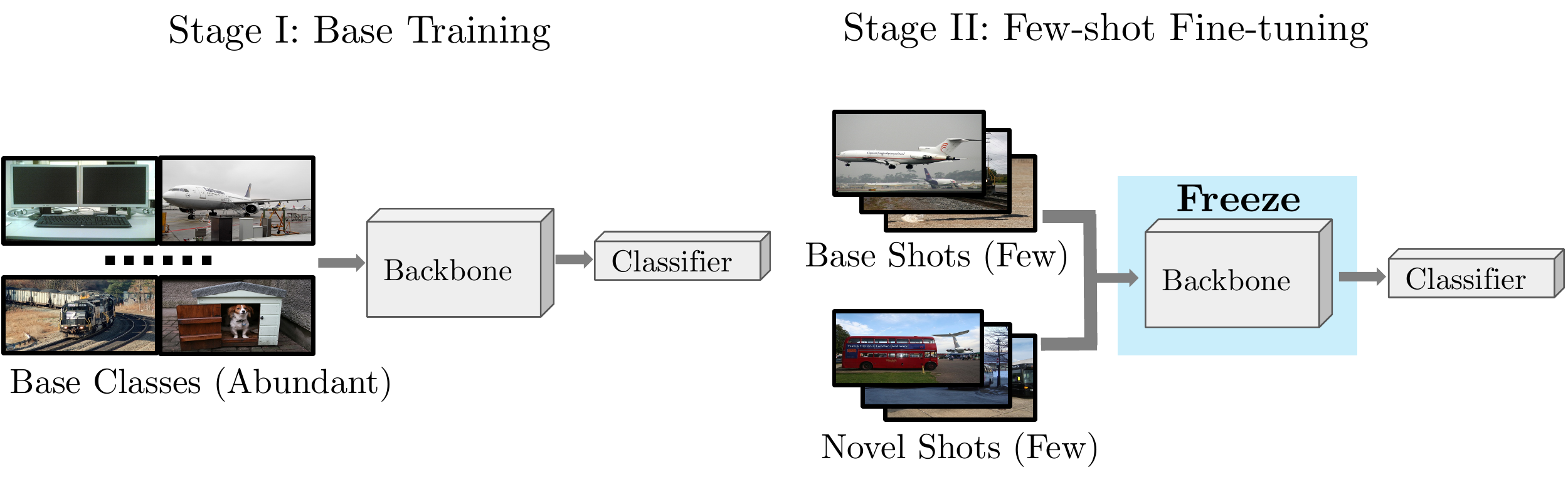}
\vspace{-0.5em}
\caption{Our proposed framework for generalized few-shot semantic segmentation. It is trained in two stages: a base training stage and a few-shot fine-tuning stage. During base training, all parameters in the network are learned using the abundance of labelled data available for base categories. In the fine-tuning stage, all layers before the classifier are frozen and the model parameters in the classifier are updated by learning from a scarcity of labelled data representing both the novel and base categories.}
\label{fig:method}
\end{figure}

The top-performing generalized few-shot segmentation approach follows the status of most few-shot segmentation approaches by using a meta-learning based approach.  Yet, observing the performance of this approach~\cite{tian2020generalized}, it performs poorly and suffers from performance saturation as the number of shots increase. For example, all reported results for GFS-Seg\cite{tian2020generalized} are below a 60\% mean Intersection over Union (\emph{mIoU}) and only a 0.3 percentage point boost is observed in total \emph{mIoU} when going from 1 to 10 shots on PASCAL-$5^i$\cite{shaban2017one}. 

Inspired by the recent success of fine-tuning for other few-shot learning tasks~\cite{wang2020frustratingly,nakamura_rev_ft}, we sought to examine whether fine-tuning methods could be successful for generalized few-shot semantic segmentation.  To our knowledge, our work is the first to introduce a fine-tuning approach.  An overview of our approach is illustrated in Figure~\ref{fig:method}.  We first demonstrate that simple fine-tuning, without any bells and whistles, achieves large improvements over prior work~\cite{tian2020generalized} on two datasets for multiple few-shot scenarios (i.e., 1, 5, and 10 shots).  Next, in the spirit of generalized learning trying to achieve strong performance on \emph{both} base and novel classes, we demonstrate that augmenting contrastive learning, using a triplet loss, effectively redistributes the performance between base and novel classes so that there is a smaller gap between them. Finally, we extend prior work by investigating how many layers should be fine-tuned.  We find that prior work's approach~\cite{wang2020frustratingly} does not generalize across tasks; i.e., optimal results are achieved by fine-tuning only the final layer for object detection~\cite{wang2020frustratingly} while optimal results are achieved fine-tuning all layers after the backbone for semantic segmentation.  In other words, the optimal feature representation to fine-tune differs across tasks.

%% file: related_work.tex
\section{Related Work}

\paragraph{Semantic Segmentation.} 
Since FCN~\cite{long2015fully} was introduced in 2015, deep convolutional neural networks have been the dominant solution for semantic segmentation. While a variety of architectures and features have been introduced to improve the FCN framework~\cite{chen2018encoder,zhao2017pyramid,chen2017rethinking,yuan2018ocnet,liu2015parsenet,yang2018denseaspp,zhao2018psanet,fu2019dual,fourure2017gridnet,sun2019high}, a typical underlying assumption is that a large number of densely annotated images is available for training. Unlike such works, we focus on the few-shot scenario where there are limited annotations for some of the object categories.

\paragraph{Few-Shot Semantic Segmentation.}
Few-shot learning was introduced for semantic segmentation in 2017~\cite{shaban2017one}.  Since, most proposed approaches employ either prototypical networks~\cite{dong2018few,wang2019panet,yang2020prototype,liu2020part,snell2017prototypical}, metric learning~\cite{rakelly2018conditional,hu2019attention,zhang2018sg,zhang2019canet,zhang2019pyramid,nguyen2019feature,wang2020few,gairola2020simpropnet,liu2020crnet,tian2020pfenet} or weight imprinting~\cite{siam2019amp}. To our knowledge, only one approach uses fine-tuning: RePRI~\cite{boudiaf2021fewshot}. However, like meta-learning, RePRI~\cite{boudiaf2021fewshot} also requires support feature maps at inference time. 

Regardless of the approach, a commonality across all of them is that at design time they were evaluated only on the novel classes.  When instead evaluating the state-of-the-art methods~\cite{zhang2019canet,tian2020pfenet,wang2019panet} in the generalized setting, where knowledge of base classes should be retained, they perform worse compared to prior work~\cite{tian2020generalized} and so, by extension, our approach (since our approach is the new state-of-the-art in the generalized setting).\footnote{While a new approach has been published since the timing of that analysis~\cite{min2021hypercorrelation}, it is unsuitable for evaluation in the generalized setting by design since it is based on multi-scaled features to predict a binary mask rather than prototypes.}, \footnote{For completeness, we also examine our approach's performance compared to the non-generalized few-shot semantic segmentation methods~\cite{zhang2019canet,tian2020pfenet,wang2019panet} when evaluating only on novel classes.  Results are reported in the Supplementary Materials. In summary, given that our method is designed to retain knowledge about base categories, it performs less well for novel-only evaluation.} 

\paragraph{Generalized Few-Shot Semantic Segmentation.}
As is common in the few-shot learning setting, the current state-of-the-art approach for the generalized setting employs a meta-learning approach~\cite{tian2020generalized}. We will show in our experiments that our fine-tuning approach outperforms this baseline~\cite{tian2020generalized} by significant margins across two datasets while not saturating in performance when observing more examples (i.e., 1, 5 and 10 shots).

\paragraph{Fine-tuning for Few-Shot Learning.}
Fine-tuning has outperformed meta-learning by large margins in few-shot learning for object detection~\cite{wang2020frustratingly} and image classification~\cite{tian2020rethinking,dhillon2019baseline}, achieving state-of-the-art performance. Our experiments reinforce the advantage of fine-tuning by showing that our fine-tuning methods achieve state-of-the-art results for few-shot semantic segmentation across two datasets. Our work also complements prior work by providing the first investigation into what are the optimal number of layers to fine-tune across different tasks.  When comparing the number of optimal layers for the two localization tasks of semantic segmentation and object detection, we found the optimal number of layers to fine-tune differs.  Our findings offer promising evidence that different feature representations are needed as a foundation for fine-tuning for different tasks. 

\paragraph{Contrastive Learning.} Contrastive learning has been a successful auxiliary task for the few-shot paradigm~\cite{Li2020FewShotIC,chen_Yao_Zhou_Dong_Zhang_2021,liu2021fewshot,Yue_2021_CVPR,LiRemote2021,chen2021scnet,Sun_2021_CVPR,majumder2021supervised,Liu_Fu_Xu_Yang_Li_Wang_Zhang_2021,Ouali_Hudelot_Tami_2021} as well as for segmentation tasks~\cite{Hu_2021_ICCV,BMVC2016_119} independently. Yet, contrastive learning has not been investigated for generalized few-shot semantic segmentation. Most similar to our work is the use of contrastive learning for few-shot object detection~\cite{wang2020frustratingly}.  While we observe that prior work's~\cite{wang2020frustratingly} contrastive learning technique (i.e., a cosine similarity) can lead to a performance drop in our setting, we find that using a triplet loss instead results in the desired outcome of an improved balance of performance between the novel and base categories.

%% file: method.tex
\section{Method}

We now introduce our few-shot semantic segmentation method. We introduce the problem definition in Section~\ref{sec:prob_def}, our two-stage fine-tuning approach in Section~\ref{sec:fine_tuning}, and auxiliary task of triplet loss in Section~\ref{sec: triplet}.  We will publicly-share all code upon publication to ensure reproducibility.
\vspace{-.5em}
\subsection{Problem Definition}
\label{sec:prob_def}
Let $\mathcal{D}_{\text{train}}$ and $\mathcal{D}_{\text{test}}$ denote the training and testing image sets of a semantic segmentation dataset respectively. In $\mathcal{D}_{\text{train}}$, let $\mathcal{C}_{b}$ denote a set of base classes that have many annotated examples and $\mathcal{C}_{n}$ represent a set of novel classes that have only a few annotated examples. For each image $I\in\mathcal{D}_{\text{test}}$, our goal is to produce a label $c_{i,j}\in \mathcal{C}_{b} \cup  \mathcal{C}_{n}$ for each 2D location $(i,j)$ of image $I$. 
\vspace{-.5em}
\subsection{Vanilla Fine-tuning Approach}
\label{sec:fine_tuning}
\paragraph{Architecture:} An overview of our architecture is shown in Figure~\ref{fig:method}.  It consists of two components: a \emph{backbone} and \emph{classifier}. To enable fair comparison with the existing generalized few-shot segmentation approach~\cite{tian2020generalized}, we design the backbone and classifier using the same architectural elements employed in that meta-learning approach: PSPNet~\cite{zhao2017pyramid} with a backbone of ResNet-50~\cite{he2016deep} up to stage 4.  For our \emph{classifier}, we use all the layers of PSPNet after stage 4 of ResNet-50.  This classifier consists of a pyramid pooling module~\cite{zhao2017pyramid} followed by a convolution layer (with 512 filters), BatchNorm layer, ReLU activation function, and a final convolution layer (with the number of filters set to match the number of classes to be predicted).  Finally, bilinear upsampling is used to match the output feature's spatial dimensions to those of the input.

\paragraph{Training:}
Our training strategy is to separate the backbone learning and classifier learning into two stages: a base training stage and a few-shot fine-tuning stage.  The aim is to first teach the backbone a feature representation that can generalize to a broader range of classes than observed during base training and then teach the classifier to use this feature representation to segment the broader range of classes when observing only a few examples per class. 

\emph{Stage I: Base Training.}
We first train both the backbone and classifier on base classes, for which there is an abundance of annotated examples.  Following prior work~\cite{zhao2017pyramid}, we train with the following loss function:
\begin{equation}
\label{eq: base_loss}
L = L_{main} + \lambda_{aux}L_{aux} .
\end{equation}
\noindent
where $L_{main}$ is the cross entropy loss for the final semantic segmentation output, and $L_{aux}$ is an auxiliary cross entropy loss for another additional classifier applied inside the backbone. As reported in PSPNet~\cite{zhao2017pyramid}, the auxiliary loss should help optimize the learning process. Following prior work~\cite{zhao2017pyramid}, we set $\lambda_{aux}=0.4$.  We use SGD as our optimizer with a learning rate of $.01$, learning rate decay of $.00001$, and momentum of $0.9$.  We use a batch size of 16 and train for 50 epochs.


\emph{Stage II: Fine-Tuning.}
We next freeze the backbone and fine-tune the classifier.\footnote{In preliminary experiments, we observed that fine-tuning the backbone led to worse results due to overfitting to the small fine-tuning set.} Our loss function, $L$, for this second stage of training is:
\begin{equation}
\label{eq: ft_loss}
L = L_{main} .
\end{equation}

\noindent
Note that, compared to the base training stage, we omit the auxiliary loss.  That is because the base weights are no longer updated.  

For training, we use the same batch size, optimizer, and learning rate as in \emph{Stage I}. Unlike \emph{Stage I}, we use as our training data a random sample of $K$ images for each base and novel class, where $K$ is the desired number of shots. The motivation for including base classes is to prevent the model from forgetting the knowledge of base classes learned in the first training stage. This sampling approach was shown to be successful for a fine-tuning approach proposed for few-shot object detection~\cite{wang2020frustratingly}.  We train for $1000$ epochs and then use the model with the best total \emph{mIoU}.
\vspace{-.5em}
\subsection{Training with an Augmented Triplet Loss}
\label{sec: triplet}
We next propose triplet loss as a form of regularization. Generally, it takes in an anchor sample ($a$), a positive sample ($p$), and a negative sample ($n$) and then aims to pull the anchor and positive points close together in feature space while pushing the anchor and negative points away from each other up to a specified margin, $\mu$.  More formally, it uses the following loss function from \cite{BMVC2016_119}:
\begin{equation}
\label{eq: triplet}
L_{triplet}(a,p,n) = max\{0,\; \delta (a,p) - \delta (a,n) + \mu \} .
\end{equation}

\noindent
where $\delta (\cdot ) = ||x_i - y_i||_2$ and $||\cdot ||_2$ is the $\ell_2$ norm. Intuitively, because triplet loss implicitly includes a notion of similarity and difference, it should help base and novel classes to become more separable in the feature space.

We experiment with using triplet loss for (a) both stages of training and (b) only for the second fine-tuning stage.  We apply triplet loss to the penultimate features of the network, $\mathcal{F}$, since they have the highest semantic resolution before the output layer and that layer has a fixed dimensionality across all datasets (unlike the last layer, which has a dimensionality that is determined by the number of classes relevant for each dataset).  Formally, let $\mathcal{C}=\mathcal{C}_b \cup \mathcal{C}_n$  and $\mathcal{T}_c, \; \forall_c \in \mathcal{C}$, denote the set of triplets extracted during a training epoch.  Since the amount of possible triplets for each class is large, for latency reasons we only sample $min( |\mathcal{T}_c| ,\; \tau)$ triplets for each class.\footnote{We set $\tau = 50$ and $\mu=1$ for our experiments.}   Let $\mathcal{C}^+$ be the set of all points of the same arbitrary class, and let $\mathcal{C}^-$ be the set of all points from other classes such that $\mathcal{C}^+ \cap \mathcal{C}^- = \emptyset$. To construct $\mathcal{T}_c$, we first randomly sample $\tau$ points from $\mathcal{C}^+$ for our anchor points, $\tau$ points from $C^+$ that are disjoint from our anchor points to act as positive examples, and $\tau$ points from $\mathcal{C}^-$ for our negative points. We then create $\mathcal{T}_c$ by randomly pairing anchor, positive, and negative points without replacement.  

\emph{Stage I: Base Training with Triplet Loss.} When augmenting triplet loss to our base training, we arrive at the following loss function:
\begin{equation}
\label{eq: trip_loss_base}
L = L_{main} + \lambda_{aux}L_{aux} + \lambda_{triplet}L_{triplet} .
\end{equation}
\noindent
where $L_{main}$ and $L_{aux}$ are the same as in Equation~\ref{eq: base_loss}, while $L_{triplet}$ from Equation~\ref{eq: triplet} is the triplet loss applied to the penultimate layer of the network.


\emph{Stage II: Fine-Tuning with Triplet Loss.}
When augmenting triplet loss for fine-tuning, we arrive at the following loss function:

\begin{equation}
\label{eq: trip_loss_ft}
L = L_{main} + \lambda_{triplet}L_{triplet} .
\end{equation}

\noindent
Given that at the start of fine-tuning our feature space is fit to base classes, we assign a larger weight to triplet loss in order to enforce the notions of similarity and difference. Specifically, by doing so, we prioritize that feature vectors of the same class should be similar and feature vectors from different classes should be dissimilar. We provide an analysis of the sensitivity of our results to changes in $\lambda_{triplet}$ and $\lambda_{base}$ in the Supplementary Materials.

%% file: experiments.tex
\section{Experiments}
\label{sec:experiments}
\vspace{-.3em}
We now evaluate the performance of our fine-tuning approaches for generalized few-shot segmentation.  

We conduct experiments using two datasets: PASCAL-$5^i$~\cite{shaban2017one} and COCO-$20^i$~\cite{shaban2017one}.   For both datasets, the few-shot scenario is mimicked by reserving a subset of classes, called folds, to act as novel classes.  PASCAL-$5^i$~\cite{shaban2017one} is built from PASCAL VOC 2012~\cite{everingham2015pascal} and we follow the dataset split in \cite{shaban2017one} to evenly split 20 object categories into four folds. The set of class indexes contained in fold $i$ is written as $\{5i+j\}$ where $i\in\{0,1,2,3\}$, $j\in\{1,2,3,4,5\}$.  COCO-$20^i$~\cite{shaban2017one} is built from MSCOCO~\cite{lin2014microsoft} and we follow the dataset split in \cite{tian2020pfenet} to evenly split 80 object categories into four folds, each fold with 20 categories. The set of class indices contained in fold $i$ is written as $\{4j-3+i\}$ where $j\in\{1,2,\dots,20\}$, $i\in\{0,1,2,3\}$.  For both datasets, we perform cross-validation, treating background class and three folds as base classes while deeming the remaining fold as novel classes. At test time, we evaluate with all the images in the validation set for evaluation.  The few-shot setting is mimicked by sampling a subset of the annotated images for the ``novel" classes.

As is standard in the literature, we evaluate performance using the Intersection-over-Union (\emph{IoU}) for each class and then average the \emph{IoU} over all relevant classes to obtain mean Intersection-over-Union (\emph{mIoU}). Final performance scores are computed by averaging \emph{mIoU} over all the folds in cross validation.  We evaluate with respect to all the classes as well as for the base and novel classes separately. 

\subsection{Analysis of Our Vanilla Fine-Tuning Approach}
We first evaluate our vanilla, two-stage fine-tuning approach described in Section~\ref{sec:fine_tuning}, which we refer to as \emph{Ours-Vanilla}.  This analysis underscores the base performance of a simple fine-tuning approach, without any bells and whistles. 
 
\paragraph{Experimental Design:} 
We evaluate performance for three common few-shot settings: 1, 5, and 10 shots. While we did not vary our learning rate (i.e., $.01$) in our experiments, for completeness we provide an analysis of the sensitivity of our method to changes in learning rate in the Supplementary Materials. In summary, we find that there is little variation in results for learning rates of $0.1$ to $0.001$, but results get worse when learning rates get even smaller.

\paragraph{Baseline:} 
For comparison, we evaluate the state-of-the-art method in generalized few-shot semantic segmentation: GFS-Seg~\cite{tian2020generalized}.  Recall, this is a meta-learning approach and is the only method benchmarked for the generalized setting. We report numbers directly from their paper since the authors did not publicly-share their code or respond to our email requests for the code.   Of note, GFS-Seg~\cite{tian2020generalized} has released three versions of their work, with the most recent work not including 10-shot results. We report their 1 and 5-shot results from the most recent version and 10-shot results from the prior version where the scores are available.

\begin{table}[!b]
\centering
\caption{Performance (\emph{mIoU}) of our models and GFS-Seg~\cite{tian2020generalized} on PASCAL-$5^i$, overall as well as with respect to the base and novel classes separately. Results are shown for our models that fine-tune different numbers of layers in the classifier (i.e., \emph{Ours-Vanilla} and \emph{Ours-ObjDetFT}) and models that augment triplet loss (i.e., all models starting with \emph{Ours-Trip}).  We show in bold the top-performing approach for each shot.  Our models outperform GFS-Seg for 5 and 10 shots and perform comparably for 1-shot.  Moreover, overall, using triplet loss leads to a slight boost to total performance with a redistribution between performance on novel and base classes that reduces their gap.}
\begin{threeparttable}
    \resizebox{\textwidth}{!}{\begin{tabular}{c c c c c c c c c c }
    \toprule
       \multirow{3}{*}{\bf\shortstack[c]{ \\\\Method}} & \multicolumn{3}{c}{\bf 1-Shot} & \multicolumn{3}{c}{\bf 5-Shot} & \multicolumn{3}{c}{\bf 10-Shot} \\
     \cmidrule(lr){2-4}
     \cmidrule(lr){5-7}
     \cmidrule{8-10}
        & \bf Base & \bf Novel & \bf Total &  \bf Base & \bf Novel & \bf Total & \bf Base & \bf Novel & \bf Total   \\
     \midrule
   GFS-Seg\cite{tian2020generalized} & 65.48 & 18.85 & 54.38 &
    66.14 & 22.41 & 55.72
    & 64.52 & 23.19 & 54.68 \\ 
    \emph{Ours-Vanilla} & 66.84 & 18.82 & 55.41 & 
    \textbf{72.03} & 46.40 & 65.93 & 
    \textbf{73.02} & 52.55 & 68.14   \\ 
    \emph{Ours-ObjDetFT} & \textbf{68.89} & 18.96 & 57.01 
    & 71.72 & 39.20 & 63.98 & 
    72.75 & 46.40 & 66.45   \\ \hdashline
    \emph{Ours-Cosine} & 68.64 & \textbf{20.26} & \textbf{57.12} & 
    71.74 & 42.96 & 64.89 &
    72.91 & 50.33 & 62.75 \\
    \emph{Ours-TripletFT} & 65.46 & 18.87 & 54.36 & 
    70.66 & 44.47 & 64.41 & 
    72.06 & 53.56 & 67.66   \\ 
    \emph{Ours-TripletAll} & 66.41 & 19.71 & 55.31 & 
    71.31 & \textbf{50.46} & \textbf{66.35} & 
    72.87 & \textbf{57.00} & \textbf{69.10}  \\
    \bottomrule\\
    \end{tabular}}
    \end{threeparttable}
\label{table:obj_generalized_pascal}
\end{table}

\paragraph{Generalized Few-shot Semantic Segmentation Results:}  
Results comparing our method to the baseline are reported in Tables \ref{table:obj_generalized_pascal} and \ref{table:obj_generalized_coco} for PASCAL-$5^i$ and COCO-$20^i$ respectively.  Overall, our approach either outperforms the baseline by considerable margins or performs comparably.

\begin{table}[!t]
  \begin{center}
  \caption{Performance (\emph{mIoU}) of our approach compared to GFS-Seg~\cite{tian2020generalized}, overall as well as with respect to the base and novel classes separately for COCO-$20^i$. Results are shown for our models that fine-tune different numbers of layers in the classifier (i.e., \emph{Ours-Vanilla} and \emph{Ours-ObjDetFT}) and models that augment triplet loss (i.e., all models starting with \emph{Ours-Trip}). We show in bold the top-performing approach for each shot.  Our approach outperforms GFS-Seg~\cite{tian2020generalized} for 5 and 10 shots while performing comparably for 1-shot.  Additionally, triplet loss results in a performance redistribution between novel and base classes that reduces the gap between them. } 
  \begin{threeparttable}
    \resizebox{\textwidth}{!}{\begin{tabular}{c c c c c c c c c c }
    \toprule
       \multirow{3}{*}{\bf\shortstack[c]{ \\\\Method}} & \multicolumn{3}{c}{\bf 1-Shot} & \multicolumn{3}{c}{\bf 5-Shot} & \multicolumn{3}{c}{\bf 10-Shot} \\
     \cmidrule(lr){2-4}
     \cmidrule(lr){5-7}
     \cmidrule{8-10}
        & \bf Base & \bf Novel & \bf Total &  \bf Base & \bf Novel & \bf Total & \bf Base & \bf Novel & \bf Total   \\
     \midrule
        GFS-Seg\cite{tian2020generalized} & 44.61 & 7.05 & 35.46 &
    45.24 & 11.05 & 36.80
    & 42.81 & 10.39 & 34.81 \\ 
    \emph{Ours-Vanilla} & 43.42 & 8.94 & 34.90 & 
    47.18 & 24.72 & \textbf{41.63} & 
    48.18 & 30.03 & \textbf{43.70}   \\ 
    \emph{Ours-ObjDetFT} & \textbf{46.02} & 8.28 & \textbf{36.70}
    & \textbf{47.57} & 20.59 & 40.91 & 
    \textbf{48.52} & 25.46 & 42.82   \\ 
    \hdashline
    \emph{Ours-TripletFT} & 44.06 & 7.53 & 35.04 & 
    45.62 & 22.97 & 39.94 & 
    46.65 & 28.28 & 42.12   \\ 
    \emph{Ours-TripletAll} &43.64  &  \textbf{9.23} & 35.14 & 
    46.61 & \textbf{28.84} & 41.36 & 
    46.61 & \textbf{34.49}  & 43.27  \\
    \bottomrule\\
    \end{tabular}}
    \end{threeparttable}
\label{table:obj_generalized_coco}
      \end{center}
\end{table}

The clear distinction between our method and GFS-Seg~\cite{tian2020generalized} is the ability to improve as more shots are observed.  For example, our approach outperforms GFS-Seg by the largest margins in the 10-shot setting, where the total \emph{mIoU} percentage point boost is 13.46 and 8.89 for PASCAL-$5^i$ and COCO-$20^i$ respectively.  In the 5-shot setting, we observe a slightly smaller performance gain of our approach over GFS-Seg, with the total \emph{mIoU} percentage point gain being 10.21 and 4.83 for PASCAL-$5^i$ and COCO-$20^i$ respectively.  In the 1-shot setting, we observe comparable performance to prior work across both datasets.  We offer our results as promising evidence that a fine-tuning approach is preferable to a meta-learning approach, since it performs better overall while not suffering from saturation.\footnote{Meta-learning saturation also is observed in the traditional few-shot setting.  For example, when quintupling the amount of shots from 1 to 5 for a top-performing method, HSNet~\cite{min2021hypercorrelation}, only a 4.2 percentage point increase in novel \emph{mIoU} is observed.}  While our findings do not negate the merit of meta-learning, they underscore a potential limitation of it in this generalized few-shot semantic segmentation setting.  

Our findings reveal that the greater performance gains of our approach over GFS-Seg~\cite{tian2020generalized} as more shots are available is due to our approach's ability to improve its learning of novel categories as more shots are observed.  For example, when the number of shots increase from 1 to 10 on PASCAL-$5^i$, we observe a 33.73 novel \emph{mIoU}percentage point increase for \textit{Ours-Vanilla} while the increase is only 4.34 percentage points for GFS-Seg.  The performance improvement of GFS-Seg tapers in its learning at around 5 shots and saturates after about 10 training shots, only gaining 0.78 percentage points in novel \emph{mIoU} despite doubling the training samples.  In contrast, our approach achieves a 5.15 percentage point increase. This saturation so quickly after only so few examples suggests that meta-learning based approaches may be prone to underfitting to novel classes for this task.  In contrast, \emph{Ours-Vanilla} learns steeply from the first few-shots and continues to benefit from additional training examples, yielding a 126\% novel \emph{mIoU} improvement on PASCAL-$5^i$ and a 189\% improvement on COCO-$20^i$ over GFS-Seg when observing 10 shots. 

Our approach is also able to retain base knowledge more efficiently than GFS-Seg, especially as the number of shots increase. For example, our approach outperforms GFS-Seg by 1.36, 5.89, and 8.50 base \emph{mIoU} percentage points for shots 1, 5, and 10 respectively on PASCAL-$5^i$. Additionally, our approach outperforms GFS-Seg in two of three tested shots for COCO-$20^i$ with a 1.94 and 5.37 percentage point increase in the 5 and 10-shot settings.

\begin{figure}[!b]
\centering
\includegraphics[width=\textwidth]{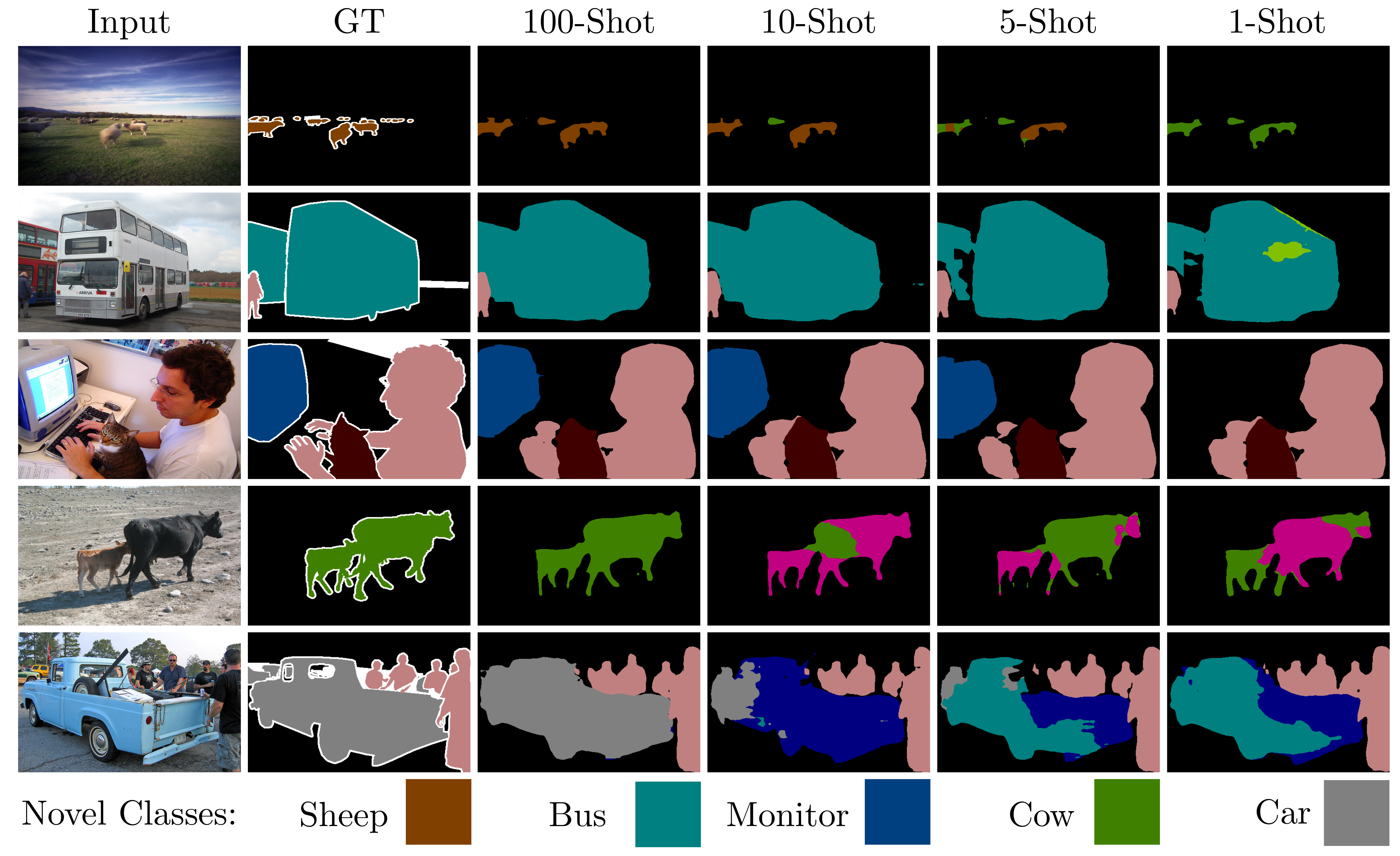}%
\caption{Comparison of results from \emph{Ours-Vanilla} for different numbers of shots. As the number of shots grows, the segmentations become more refined and novel class segmentation improves. (White in the GT indicates pixels to ignore)}
\label{fig: comp_van}
\end{figure}

We extend the above results to show the improvement of \emph{Ours-Vanilla} from 10 to 100 shots in the Supplementary Materials. We observe a further 12.32 percentage point improvement on PASCAL-$5^i$ and a further 10.20 percentage point improvement on COCO-$20^i$, again reinforcing that our approach keeps learning as we increase the number of training samples. We show qualitative results for intervals between 1 and 100 shots for \emph{Ours-Vanilla} in Figure \ref{fig: comp_van}. These results exemplify how, as we increase in the number of training examples, we see more fine-grained segmentations and better novel class recognition. For example, as shown in the last row, as we increase the amount of available shots we observe that the car is more appropriately segmented into the correct novel class.

\hfill%

\subsection{Analysis of Different Fine-Tuning Approaches}
\label{sec: ft_num_layers}
We next analyze how the number of layers that are fine-tuned impacts performance.  We compare our approach of fine-tuning all layers after the backbone with fine-tuning only the last convolutional layer, as done for the few-shot object detection approach~\cite{wang2020frustratingly}.  Accordingly, we refer to the latter of fine-tuning only the last convolutional layer as \emph{Ours-ObjDetFT}. 
\vspace{-1em}
\paragraph{Results:} Results are shown in Tables \ref{table:obj_generalized_pascal} and \ref{table:obj_generalized_coco} for PASCAL-$5^i$ and COCO-$20^i$ respectively.  

Compared to the state-of-the-art (meta-learning) baseline~\cite{tian2020generalized}, we observe that fine-tuning only the final convolutional layer leads to improved or comparable performance.  This is the case overall as well as with respect to base and novel classes separately.  Consequently, fine-tuning is an effective strategy across a wide range of learnable parameters; e.g., 0.08\% are fine-tuned for the final convolutional layer (i.e., \emph{Ours-ObjDetFT}) compared to 47.05\% are fine-tuned for the entire classifier backbone (i.e., \emph{Ours-Vanilla}).  Altogether, these findings reinforce our argument for fine-tuning as a preferred solution over meta-learning for generalized few-shot semantic segmentation.

\begin{figure}[!b]
\centering
\includegraphics[width=\textwidth]{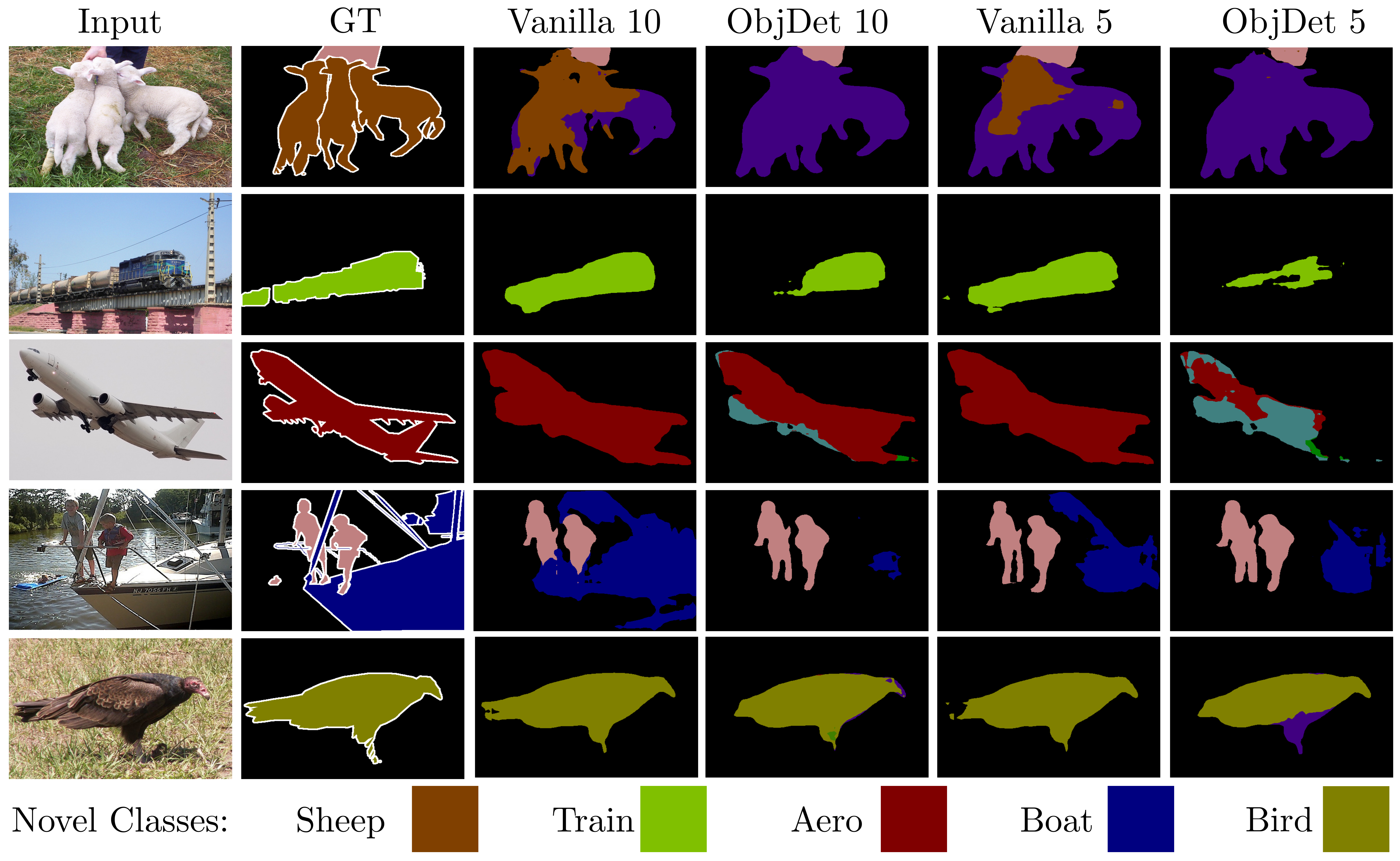}%
\caption{Results for 5 and 10 shots from \emph{Ours-Vanilla} and \emph{Ours-ObjDetFT}. We observe slightly better novel class identification and segmentation boundaries when fine-tuning more layers (i.e., \emph{Ours-Vanilla}). (White in the GT indicates pixels to ignore)}%
\label{fig: comp_510}
\end{figure}\hfill%

Our results also offer initial insights into which features are most useful to fine-tune. We consistently observe better total \emph{mIoU} when fine-tuning more layers for the 5-shot and 10-shot settings.  Inspecting the performance on the base and novel classes independently, we observe across both datasets that the advantage of fine-tuning more layers is that the performance on the novel categories improves considerably.  Our qualitative results in Figure \ref{fig: comp_510} complement these findings. Specifically, fine-tuning with more layers leads to slightly better novel class identification and overall segmentation quality, especially in the case when multiple classes are present. For example, in the first row we observe correct novel class identification with the sheep when fine-tuning with more layers, while only fine-tuning the last layer mis-classifies the sheep. Furthermore, in the second and fourth rows, fine-tuning with more layers more appropriately distinguishes the novel classes, train and boat, from the background compared to fine-tuning only the last layer. We suspect that a smaller number of parameters may lack sufficient representational power when trying to fine-tune them to novel classes.  However, in the 1-shot scenario, fine-tuning fewer layers is slightly better due to its ability to retain base category knowledge better.

\begin{table}[!b]
  \begin{center}
  \caption{Comparison between two fine-tuning approaches: (i) only the last layer in the network (i.e., Last) and (ii) fine-tuning all layers after the backbone (i.e., Backbone), as done by prior work for few-shot object detection\cite{wang2020frustratingly}.  Results are reported for PASCAL-$5^i$. Overall, we observe that different fine-tuning approaches are better for the two tasks.  The exception is for the 1-shot setting, where fine-tuning the last convolutional layer is consistently the optimal choice.} 
  \begin{threeparttable}
    \resizebox{\textwidth}{!}{\begin{tabular}{c c c c c c c c c c }
    \toprule
       \multirow{3}{*}{\bf\shortstack[c]{ \\\\Method}} & \multicolumn{3}{c}{\bf 1-Shot} & \multicolumn{3}{c}{\bf 5-Shot} & \multicolumn{3}{c}{\bf 10-Shot} \\
     \cmidrule(lr){2-4}
     \cmidrule(lr){5-7}
     \cmidrule{8-10}
        & \bf Base & \bf Novel & \bf Total &  \bf Base & \bf Novel & \bf Total & \bf Base & \bf Novel & \bf Total   \\
        \midrule
     \emph{Ours} & Last & Last & Last
     & Backbone & Backbone & Backbone
     & Backbone & Backbone & Backbone\\ 
     \emph{FSDet}\cite{wang2020frustratingly} & Last & Last & Last & Last & Last & Last & Last & Last & Last\\
    \bottomrule\\
    \end{tabular}}
    \end{threeparttable}
\label{table: fsdet_comp}
      \end{center}
\end{table}

Finally, we analyze whether a single fine-tuning approach generalizes across tasks, specifically object detection and semantic segmentation.  To do so, we also assess how our approach of fine-tuning all layers after the backbone impacts performance when used with the few-shot object detection approach~\cite{wang2020frustratingly}.  Results are shown for PASCAL-$5^i$ in Table \ref{table: fsdet_comp}.  We observe that neither tested fine-tuning approach is optimal for the two tasks.  Overall, object detection performs \emph{worse} when fine-tuning all layers after the backbone whereas semantic segmentation performs \emph{better} when fine-tuning all layers after the backbone.  We suspect this contrasting finding may be in part because a there is a magnitude of difference in the amount of representational power available in the last layer of the network between the two tasks. Few-shot object detection~\cite{wang2020frustratingly} has 0.26\% and 1.00\% of total model parameters available in the last convolutional layer for PASCAL-$5^i$ and COCO-$20^i$ respectively while our few-shot semantic segmentation approach has 0.02\% and 0.08\% of total model parameters in the last convolutional layer for PASCAL-$5^i$ and  COCO-$20^i$ respectively .  Another possible reason for the different tasks performing better with different fine-tuning techniques is that they have different requirements.  Intuitively, object detection may need to fine-tune fewer parameters since it produces fewer predictions overall (i.e., bounding box and classification for each object) than semantic segmentation (i.e., per-pixel predictions). 


\subsection{Analysis of Augmenting Triplet Loss}
We next examine the impact of augmenting triplet loss to our baseline approach (i.e., \emph{Ours-Vanilla}), both when using it only for fine-tuning at the second stage (i.e., \emph{Ours-TripletFT}) as well as for both stages (i.e., \emph{Ours-TripletAll}).  

\paragraph{Baseline:}  For comparison, we also evaluate substituting triplet loss with the contrastive learning approach used by prior work~\cite{wang2020frustratingly} for few-shot object detection:  cosine similarity.  We refer to this variant as \emph{Ours-Cosine}.  For efficiency, we evaluate only on the smaller dataset, PASCAL-$5^i$.  

\vspace{-0.25em}
\paragraph{Results:} Results are reported in Tables \ref{table:obj_generalized_pascal} and \ref{table:obj_generalized_coco} for PASCAL-$5^i$ and COCO-$20^i$ respectively.  Overall, we observe that augmenting triplet loss for both stages (i.e., \emph{Ours-TripletAll}) outperforms augmenting triplet loss only for the fine-tuning stage (i.e., \emph{Ours-TripletFT}).  Therefore, we focus on this variant for our subsequent analysis.

\begin{figure}[!t]
    \centering
    \includegraphics[scale=.18]{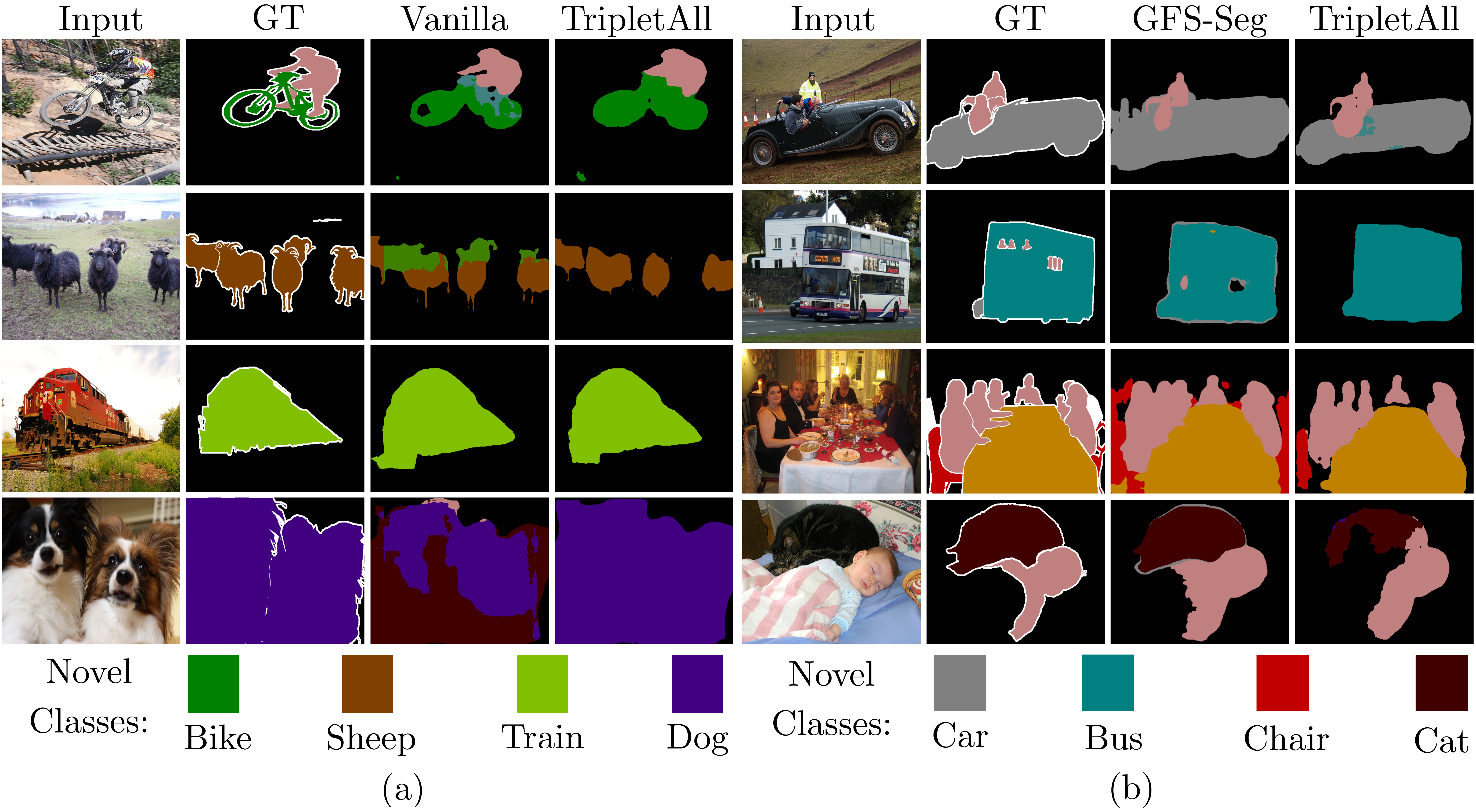}
    \caption{(a) Comparison of \emph{Ours-Vanilla} and \emph{Ours-TripletAll}. Overall, the addition of triplet loss helps improve the novel class segmentations for both single object and multi-object scenarios. (b) Comparison of results from our method to those in GFS-Seg~\cite{tian2020generalized}, where we leverage results from the GFS-Seg~\cite{tian2020generalized} paper to enable comparison (since that code base is not publicly-available to support further comparisons). From left to right: input image, ground truth segmentation of base and novel classes, results from GFS-Seg~\cite{tian2020generalized}, and then results from \emph{Ours-TripletAll}. The novel classes are: car, bus, chair, and cat with white meaning ignore those pixels. The first three rows exemplify the advantage of our approach over GFS-Seg, while the last row demonstrates a failure case for our method.}
    \label{fig: subfig}
\end{figure}

Compared to our baseline approach that lacks triplet loss (i.e., \emph{Ours-Vanilla}), augmenting triplet loss (i.e., \emph{Ours-TripletAll}) performs either comparably or slightly better while redistributing the performance between the base and novel classes so that the gap is smaller.  In particular, across all tested shots (i.e., 1, 5, and 10) for both datasets, we observe the performance on novel categories improves while the performance on base categories drops slightly.  Qualitative results comparing the two implementations that use and lack triplet loss are shown in Figure~\ref{fig: subfig}a. These examples illustrate that augmenting triplet loss more often correctly identifies novel classes while sometimes producing finer-grained segmentations. As exemplified in the second row, despite arriving at more correct class predictions (e.g., correctly predicting ``sheep"), triplet loss can sometimes result in inferior segmentation results compared to the model that lacks triplet loss.  While the overall quantitative trend from these results is that triplet loss helps create a more generalizable feature space that more effectively can separate novel classes, our qualitative results highlight that there is still room for improvement to consistently generate more fine-grained segmentations. 

We also compare our model's confidence in predicting novel classes for when it augments triplet loss (i.e., \emph{Ours-TripletAll}) versus lacks triplet loss (i.e., \emph{Ours-Vanilla}).  To do so, we examine the softmax probabilities of novel class predictions for a random subset (i.e., $2\text{e}6$ points) of predictions which match the ground truth across all shots.  While complete results are shown in the Supplementary Materials, our key observation is that triplet loss results in more confident predictions. For instance, the mean confidence for sampled points on PASCAL-$5^i$ with triplet loss is 41.94\%, while the mean confidence without triplet loss is 32.44\%. For COCO-$20^i$ the mean confidence when using triplet loss is 48.14\%, while the mean confidence without it is 40.69\%. 

We next provide qualitative results to exemplify how our more balanced fine-tuning approach (i.e., \emph{Ours-TripletAll}) compares to the prior state-of-the-art method, GFS-Seg~\cite{tian2020generalized}. Since prior work~\cite{tian2020generalized} has not published their code at the time of submission, we focus only on examples that those authors provided in their paper.  Results are shown in Figure~\ref{fig: subfig}b.  In the first example (i.e., row 1), we observe that our approach generates a better segmentation of the car (i.e., the novel class) and a better gap between the person's arm and the car.  More generally, this suggests that our approach may be better able to distinguish novel classes from the background class while better capturing fine-grained boundary details.  The second example (i.e., row 2) shows that neither \emph{Ours-TripletAll} nor GFS-Seg~\cite{tian2020generalized} are able to segment the car and people from the bus (i.e., the novel class), but our approach is able to more appropriately segment the bus by not introducing holes to the segmentation. The third example (i.e., row 3) shows that \emph{Ours-TripletAll} again avoids mistakenly segmenting the background class for the novel class of chair as is done by GFS-Seg~\cite{tian2020generalized}. The final example (i.e., row 4) shows a failure case of \emph{Ours-TripletAll} compared to GFS-Seg~\cite{tian2020generalized}.  As shown, our approach is not able to segment the cat (i.e., the novel class) as well and confuses some of the cat with the person class, highlighting that while globally our approach outperforms GFS-Seg\cite{tian2020generalized}, local failure cases still occur.

We also assess how prior work's contrastive baseline of cosine similarity (i.e., \emph{Ours-Cosine}) used for few-shot object detection performs for our generalized few-shot semantic segmentation task.  Overall, it not only leads to worse results compared to both of our triplet loss approaches, but also to our vanilla fine-tuning approach (i.e., \emph{Ours-Vanilla}).  For example, we observe a performance drop of 1.04 and 5.39 percentage points in total \emph{mIoU} in the 5 and 10-shot settings respectively.  These performance drops stem from a fall in performance on novel categories.  Our findings hint that providing \emph{explicit} (i.e., supervised) knowledge to our model about the similarities and differences between pairs of pixels by computing a triplet loss is preferable to \emph{implicitly} (i.e., unsupervised) contrasting each class from each other with cosine similarity.  

For completeness, we report in the Supplementary Materials results for applying triplet loss to the second fine-tuning approach we analyze in this paper of only fine-tuning the last convolutional layer: \emph{Ours-ObjDetFT}.  Our findings reinforce the benefits we observed from augmenting triplet loss to our vanilla fine-tuning approach.  Specifically, we still observe comparable or better performance overall paired with a redistribution of performance between base and novel classes so that the gap between them becomes smaller.

We also report our findings for \emph{Ours-TripletAll's} sensitivity to perturbations in the number of novel classes present in the Supplementary Materials. In summary, we observe decreased performance when there is an increased amount of novel classes, and increased performance when the novel class amount is decreased. Both performance changes stem from changes in base class performance, suggesting that our ability to retain base knowledge while introducing novel classes may be affected by the ratio of base to novel classes presented during training (while novel categories are relatively unaffected).

%% file: conclusion.tex
\section{Conclusion}
\vspace{-1em}
We presented a simple, yet effective, two-stage fine-tuning approach for generalized few-shot semantic segmentation. We show that two different fine-tuning based approaches that fine-tune a different number of layers both can achieve new state-of-the-art results, despite their major differences in representational power.   Moreover, we observe that the benefit of fine-tuning approaches over the existing state-of-the-art baseline increases as the numbers of shots grows.  To support generalization of our findings, we demonstrate these results on two datasets across 1, 5, 10, and 100 shots. We also demonstrate that augmenting contrastive learning to our approach in the form of triplet loss, results in a desirable redistribution in performance such that the performance on novel categories increases to narrow the gap in performance between the novel and base categories. 



%% file: supplementary.tex
\hspace{-1em}\textbf{\Large Supplementary Materials}\\\\
This document supplements the main paper with the following:
\begin{enumerate}
    
    \item Hyper-parameter selection and sensitivity analysis. (supplements \textbf{Sections 3.3 and 4.1})
    \item Analysis of augmenting triplet loss. (supplements \textbf{Section 4.3})
    \item Auxiliary analysis of our approach. (supplements \textbf{Sections 4.1, 4.2, and 4.3})
    \item Analysis of different fine-tuning approaches for the related problem of object detection. (supplements \textbf{Section 4.2)}
\end{enumerate}

\section{Hyper-parameter Selection and Sensitivity Analysis}

\subsection{Influence of Learning Rate}
We analyzed how changes to the learning rate impact the performance of our approach.  To do so, we tested our baseline approach (i.e., \emph{Ours-Vanilla}) on PASCAL-$5^i$ with five learning rates evenly spaced from $0.1$ to $1e-5$. We use the same random seed throughout each experiment to keep the learning rate as the only dependent variable. As a consequence, results may differ slightly from those reported in the main paper.  Results are shown in Table \ref{table: learn}. We observe little performance change (i.e., total \emph{mIoU}) between larger learning rates (i.e., $0.1$, $0.01$, $0.001$).  However, for smaller learning rates (i.e., $1\mathrm{e}{-4}$, $1\mathrm{e}{-5}$), we observe considerably worse results. We hypothesize that smaller learning rates are too small to reach a sufficient local optima during training. 

\begin{table}[!ht]
  \begin{center}
  \caption{Sensitivity of \emph{Ours-Vanilla} to changes in learning rate. We observe similar results (with respect to total \emph{mIoU}) across larger learning rates, but results become considerably worse for smaller learning rates.} 
  \begin{threeparttable}
    \resizebox{\textwidth}{!}{\begin{tabular}{c c c c c c c c c c }
    \toprule
       \multirow{3}{*}{\bf\shortstack[c]{ \\\\Learning Rate}} & \multicolumn{3}{c}{\bf 1-Shot} & \multicolumn{3}{c}{\bf 5-Shot} & \multicolumn{3}{c}{\bf 10-Shot} \\
     \cmidrule(lr){2-4}
     \cmidrule(lr){5-7}
     \cmidrule{8-10}
        & \bf Base & \bf Novel & \bf Total &  \bf Base & \bf Novel & \bf Total & \bf Base & \bf Novel & \bf Total   \\
        \midrule
     $0.1$ & 64.42 & \bf 29.19 & 56.03 & 69.95 & 43.50 & 63.65 & 71.27 & 51.87 & 66.65\\
     $0.01$ & 67.19 & 21.80 & \bf 56.38 & \bf 71.71 & 43.14 & 64.91 & 72.41 & 50.22 & 67.13\\
     $0.001$ & \bf 67.26 & 6.03 & 52.69 & 71.64 & \bf 45.27 & \bf 65.36 & \bf 72.42 & \bf 52.39 & \bf 67.65\\
     $1\mathrm{e}{-4}$ & 67.27 & 0.00 & 51.25 & 66.95 & 4.40 & 52.06 & 70.61 & 25.51 & 59.87\\
     $1\mathrm{e}{-5}$ & 68.07 & 0.00 & 51.86 & 67.81 & 0.00 & 51.66 & 67.86 & 0.00 & 51.70\\
    \bottomrule\\
    \end{tabular}}
    \end{threeparttable}
\label{table: learn}
      \end{center}
\end{table}

\subsection{Weight of Triplet Loss in Loss Function}
We chose the first values that we tried for weighting the triplet loss in our loss function since those values already led to improvements over our vanilla approach that lacks triplet loss (i.e., \emph{Ours-Vanilla}).  Specifically, for equations 4 and 5 in the main paper, we set this hyper-parameter (i.e., $\lambda_{triplet}$) to $0.5$ for base training, and $1.0$ for fine-tuning. 

\begin{table}[!t]
  \begin{center}
  \caption{Study on the impact of the value assigned to the hyperparameter  $\lambda_{triplet}$ during both base training (i.e., BT) and fine-tuning (i.e., FT). We observe consistently better performance across all shots when using triplet loss in both stages compared to only using triplet loss in the fine-tuning stage or omitting triplet loss completely. The first row matches the approach we used in the main paper for \emph{Ours-Vanilla} while the third row matches the approach we used in the main paper for \emph{Ours-TripletAll}.} 
  \begin{threeparttable}
    \resizebox{\textwidth}{!}{\begin{tabular}{c c c c c c c c c c c c }
    \toprule
      & \multicolumn{2}{c}{$\mathbf{\lambda_{triplet}}$}  & \multicolumn{3}{c}{\bf 1-Shot} & \multicolumn{3}{c}{\bf 5-Shot} & \multicolumn{3}{c}{\bf 10-Shot} \\
     \cmidrule(lr){4-6}
     \cmidrule(lr){7-9}
     \cmidrule{10-12}
        & \bf BT & \bf FT & \bf Base & \bf Novel & \bf Total &  \bf Base & \bf Novel & \bf Total & \bf Base & \bf Novel & \bf Total   \\
        \midrule
   &  0 & 0 & 65.59 & 21.11 & 55.24 & 
    \bf 71.95 & 44.88 & 65.51 & 
    72.41 & 53.44 & 67.89  \\ 
    &  0 & 1 &  66.33 & 19.39 & 55.15 &
    70.96 & 42.56 & 64.20 &
    72.54 & 50.87 & 67.38 \\
    &  0.5 &  1 & 66.77 & 15.56 & 54.59 &
    71.46 & \bf 50.61 & \bf 66.49 &
    72.31 & \bf 57.53 & 68.79\\
    &  1 & 0.5 & \bf 67.81 & 15.05 & 55.24 &
    71.81 & 46.59 & 65.81 &
    72.74 & 56.65 & \bf 68.88\\
    &  1 &  1 & 66.31 & \bf 25.38 & \bf 56.56 &
    71.22 & 47.43 & 65.55 &
    \bf 72.75 & 53.21 & 68.10\\
    \bottomrule\\
    \end{tabular}}
    \end{threeparttable}
\label{table: trip}
      \end{center}
\end{table}


For completeness, we conducted follow-up analysis to examine how sensitive our method's performance (i.e., total \emph{mIoU}) is to changes in this hyper-parameter (i.e., $\lambda_{triplet}$).  We varied the value of $\lambda_{triplet}$ for both base and fine-tuning stages and conducted experiments on PASCAL-$5^i$. We use the same random seed throughout each experiment to keep this hyper-parameter as the only dependent variable. As a consequence, results may differ slightly from those reported in the main paper. 

Results are shown in Table \ref{table: trip}. Regardless of what weights values are given to triplet loss, we observe, across all shots, consistently better results (with respect to total \emph{mIoU}) when including triplet loss in both stages of training compared to only including triplet loss in the fine-tuning stage (i.e., \emph{Ours-TripletFT}) or omitting triplet loss completely (i.e., \emph{Ours-Vanilla}). Performance on the base classes is relatively unaffected by changes in $\lambda_{triplet}$, while performance improves for novel classes when including triplet loss in both stages rather than omitting triplet loss in one or both stages (i.e., \emph{Ours-Vanilla} and \emph{Ours-TripletFT}). 

Our findings in this study also reinforce those reported in the main paper which show that including triplet loss only in the fine-tuning stage leads to worse results compared to our baseline approach (i.e., \emph{Ours-Vanilla}). We hypothesize that trying to optimize a new objective function (i.e., cross entropy and triplet loss) with limited data leads to degradation in performance when including triplet loss only in the fine-tuning stage.

\section{Analysis of Augmenting Triplet Loss}
\subsection{Fine-grained Analysis of Performance on Base Classes}
We examine how the inclusion of triplet loss in the base stage of training affects predictive performance on base classes. To do so, we compare including triplet loss in the base stage to the baseline of not using triplet loss when training the base stage.  We then report the performance of our approach across all base categories \emph{mIoU} for each fold on two datasets (i.e., PASCAL-$5^i$ and COCO-$20^i$).  

Results are reported in Table \ref{table:base-ab-voc}. Across all folds, the \emph{base mIoU} is similar whether including versus not including triplet loss.  These results reinforce our finding in the main paper that the benefit of including a triplet loss during the first-stage of training is to support downstream generalization to novel classes (i.e., as demonstrated in Tables 1 and 2 in the main paper). We hypothesize that we do not see a significant change in base class performance after the first stage because we have a sufficient amount of training examples for base classes.

\begin{table}[!h]
\caption{Comparison of base training with and without triplet loss for both PASCAL-$5^i$ and COCO-$20^i$. The addition of triplet loss does not have a significant impact on base category performance after the first-stage of training.}
\begin{center}
\begin{tabular}{  c  c  c  c }
\toprule
\bf Method & \bf Fold & \bf \quad PASCAL-$\mathbf{5^i}$ & \bf \quad COCO-$\mathbf{20^i}$\\
\hline
Without Triplet & 0 &  \textbf{72.92} & 42.65 \\
With Triplet & 0 &  72.25 &  \textbf{43.25}\\
\hline
Without Triplet & 1 & 62.91 & 47.67\\
With Triplet & 1 & \textbf{63.62} & \textbf{48.25}\\
\hline
Without Triplet & 2 &  64.33 &  \textbf{51.25} \\
With Triplet & 2 & \textbf{65.18} & 50.61\\
\hline
Without Triplet & 3 &  73.38  &  \textbf{49.41} \\
With Triplet & 3 & \textbf{74.41} & 49.40 \\
\hline
Without Triplet & Avg. & 68.38  & 47.75\\
With Triplet & Avg. & \textbf{68.86} & \textbf{47.88} \\
\bottomrule
\end{tabular}
\end{center}
\label{table:base-ab-voc}
\end{table}
 
\subsection{Influence of Triplet Loss When Fine-Tuning the Last Layer}
We examine how the addition of triplet loss affects performance to our approach, when we fine-tune only the last layer.  We refer to this variant as  \emph{Ours-TripBaseFTLast}.  We compare this approach to the baseline of fine-tuning only the last layer without triplet loss (i.e., \emph{Ours-ObjDetFT}). To demonstrate the generalization of our findings, we conduct our experiments on two datasets (i.e., PASCAL-$5^i$ and COCO-$20^i$). 

Results are shown in Table~\ref{table: lastft}. Overall, we observe consistently higher novel and total \emph{mIoU} and comparable base \emph{mIoU} when using our variant with triplet loss (i.e., \emph{Ours-TripBaseFTLast}). This provides further evidence that triplet loss helps form a feature space that generalizes well to novel classes. 

\begin{table}[!h]
  \begin{center}
  \caption{Comparison of augmenting triplet loss to the base training stage of \emph{Ours-ObjDetFT} (i.e., \emph{Ours-TripBaseFTLast}). The top part of the table represents results on PASCAL-$5^i$, while the bottom represents results on COCO-$20^i$. Overall, we observe consistently higher novel class performance and comparable base class performance compared to \emph{Ours-ObjDetFT}.} 
  \begin{threeparttable}
    \resizebox{\textwidth}{!}{\begin{tabular}{c c c c c c c c c c }
    \toprule
       \multirow{3}{*}{\bf\shortstack[c]{ \\\\Method}} & \multicolumn{3}{c}{\bf 1-Shot} & \multicolumn{3}{c}{\bf 5-Shot} & \multicolumn{3}{c}{\bf 10-Shot} \\
     \cmidrule(lr){2-4}
     \cmidrule(lr){5-7}
     \cmidrule{8-10}
        & \bf Base & \bf Novel & \bf Total &  \bf Base & \bf Novel & \bf Total & \bf Base & \bf Novel & \bf Total   \\
        \midrule
     \emph{Ours-ObjDetFT} &  68.89 & 18.96 & 57.01 
    & 71.72 & 39.20 & 63.98 & 
    72.75 & 46.40 & 66.45 \\
    \emph{Ours-TripBaseFTLast} & \bf 68.99 & \bf 25.40 & \bf 58.61 & \bf 71.88 & \bf 47.24 & \bf 66.01 & \bf 73.19 & \bf 51.54 & \bf 68.04\\
    \midrule
    \emph{Ours-ObjDetFT} &   \bf 46.02 & 8.28 & 36.70
    & \bf 47.57 & 20.59 & 40.91 & 
   \bf 48.52 & 25.46 & 42.82 \\
   \emph{Ours-TripBaseFTLast} &  45.26 & \bf 11.62 & \bf 36.95 &  47.38 & \bf 27.87 & \bf 42.57 & 48.06 & \bf 31.15 & \bf 43.88\\
    \bottomrule\\
    \end{tabular}}
    \end{threeparttable}
\label{table: lastft}
      \end{center}
\end{table}

\section{Auxiliary Analysis of Our Approach}
\subsection{Confidence in Predictions for Novel Classes}
As described in the main paper, we examine how the inclusion of triplet loss affects the confidence  of correct novel class predictions.  We randomly sampled without replacement a subset (i.e., $2e6$ points) of predictions which match the ground truth across all shots for both our baseline approach (i.e., \emph{Ours-Vanilla}) and our baseline approach augmented with triplet loss (i.e., \emph{Ours-TripletAll}). We measure confidence as the softmax probabilities for correct novel classes. 

\begin{figure}[!hbt]
\centering
\includegraphics[width=0.7\textwidth]{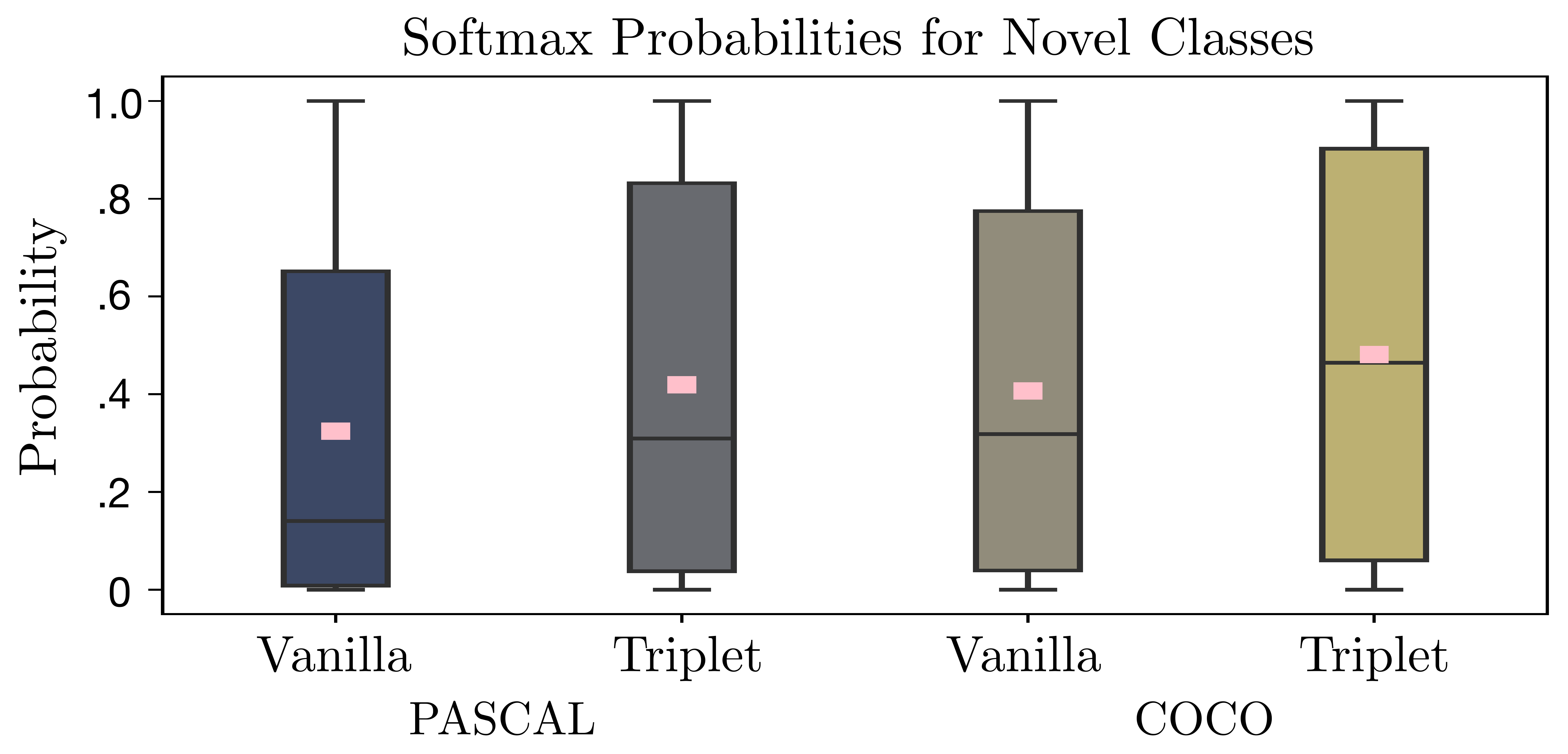}%
\caption{Boxplot showing the softmax scores from \emph{Ours-Vanilla} and \emph{Ours-TripletAll} on points sampled from activations of correct novel class predictions for both COCO-$20^i$ and PASCAL-$5^i$ validation sets.  Each box denotes the median score as the central mark, the 25th and 75th percentiles scores as the box edges, and the most extreme data points not considered outliers as the whiskers. Means are denoted with pink boxes. Overall, we observe that triplet loss leads to more confident correct predictions.}
\label{fig: boxplot}
\end{figure}

A visualization of the distribution of confidences is shown in Figure \ref{fig: boxplot} as boxpots. We observe that the mean confidence on PASCAL-$5^i$ with triplet loss is 41.94\%, while the mean confidence without triplet loss is 32.44\%. For COCO-$20^i$ the mean confidence when using triplet loss is 48.14\%, while the mean confidence without it is 40.69\%. This demonstrates that not only can we achieve better performance (i.e., total \emph{mIoU}) with the inclusion of triplet loss, but that the approach is more confident in its predictions.

\hfill%

\subsection{100-Shot Semantic Segmentation}
As described in the main paper, we also evaluated our approaches in the 100-shot setting. We compare the results when 100 shots are available to when only 10 shots are available in order to demonstrate how our approaches scale with additional training samples. 

Results are shown in Tables \ref{table: 100-pascal} and \ref{table: 100-coco} for PASCAL-$5^i$ and COCO-$20^i$ respectively. Overall, we observe significant improvements in base, novel, and total \emph{mIoU} when using 100 shots instead of 10 shots. In other words, learning improves as more shots are observed.  This finding contrasts what we observed for meta-learning.  A reason for this distinction could be that our approach directly computes a loss on novel categories, explicitly discriminating between categories, while meta-learning leverages support features which may not be informative enough to generalize well to novel classes.

\begin{table}[!h]
  \begin{center}
  \caption{Results for training our approach with 10 and 100 shots on PASCAL-$5^i$. We continue to see improvements across base, novel, and total \emph{mIoU} without saturating as the number of shots available increase.}
  \begin{threeparttable}
    \begin{tabular}{c c c c c c c}
    \toprule
       \multirow{3}{*}{\bf\shortstack[c]{ \\\\Method}} & \multicolumn{3}{c}{\bf 10-Shot} &\multicolumn{3}{c}{\bf 100-Shot} \\
     \cmidrule(lr){2-4}
     \cmidrule{5-7}
        & \bf Base & \bf Novel & \bf Total & \bf Base & \bf Novel & \bf Total   \\
        \midrule
    \emph{Ours-Vanilla} & \textbf{73.02} & 52.55 & 68.14  & 75.33 & 64.87 & 72.84\\
    \emph{Ours-ObjDetFT} & 72.75 & 46.40 & 66.45  &  74.25 & 52.70 & 69.12\\
    \hdashline
    \emph{Ours-TripletFT} & 72.06 & 53.56 & 67.66   & 74.52 & 65.18 & 72.03\\
    \emph{Ours-TripletAll} &  72.87 & \textbf{57.00} & \textbf{69.10}  & \textbf{75.37} & \textbf{68.35} & \textbf{73.70}\\
    \bottomrule\\
    \end{tabular}
    \end{threeparttable}
\label{table: 100-pascal}
      \end{center}
\end{table}

\begin{table}[!h]
  \begin{center}
  \caption{Results for training our approach with 10 and 100 shots on COCO-$20^i$. We continue to see improvements across base, novel, and total \emph{mIoU} without saturating as the number of shots available increase.} 
  \begin{threeparttable}
    \begin{tabular}{c c c c c c c}
    \toprule
       \multirow{3}{*}{\bf\shortstack[c]{ \\\\Method}} & \multicolumn{3}{c}{\bf 10-Shot} &\multicolumn{3}{c}{\bf 100-Shot} \\
     \cmidrule(lr){2-4}
     \cmidrule{5-7}
        & \bf Base & \bf Novel & \bf Total & \bf Base & \bf Novel & \bf Total   \\
    \midrule
        \emph{Ours-Vanilla} &  48.18 & 30.03 & \textbf{43.70} & \textbf{50.94} & 40.23 & 48.30\\
        \emph{Ours-ObjDetFT} & \textbf{48.52} & 25.46 & 42.82 & 50.14 & 32.04 & 45.67\\
        \hdashline
        \emph{Ours-TripletFT} &  46.65 & 28.28 & 42.12 & 50.93 & 40.71 & 48.41\\
        \emph{Ours-TripletAll} &  46.61 & \textbf{34.49}  & 43.27 &  50.01 & \textbf{45.19} & \textbf{48.81}\\
    \bottomrule\\
    \end{tabular}
    \end{threeparttable}
\label{table: 100-coco}
      \end{center}
\end{table}

\clearpage 
\subsection{Qualitative Results}
We provide additional qualitative results from our approaches for both datasets (i.e., COCO-$20^i$ and PASCAL-$5^i$).

\begin{figure*}[!h]
\centering
\includegraphics[width=\textwidth]{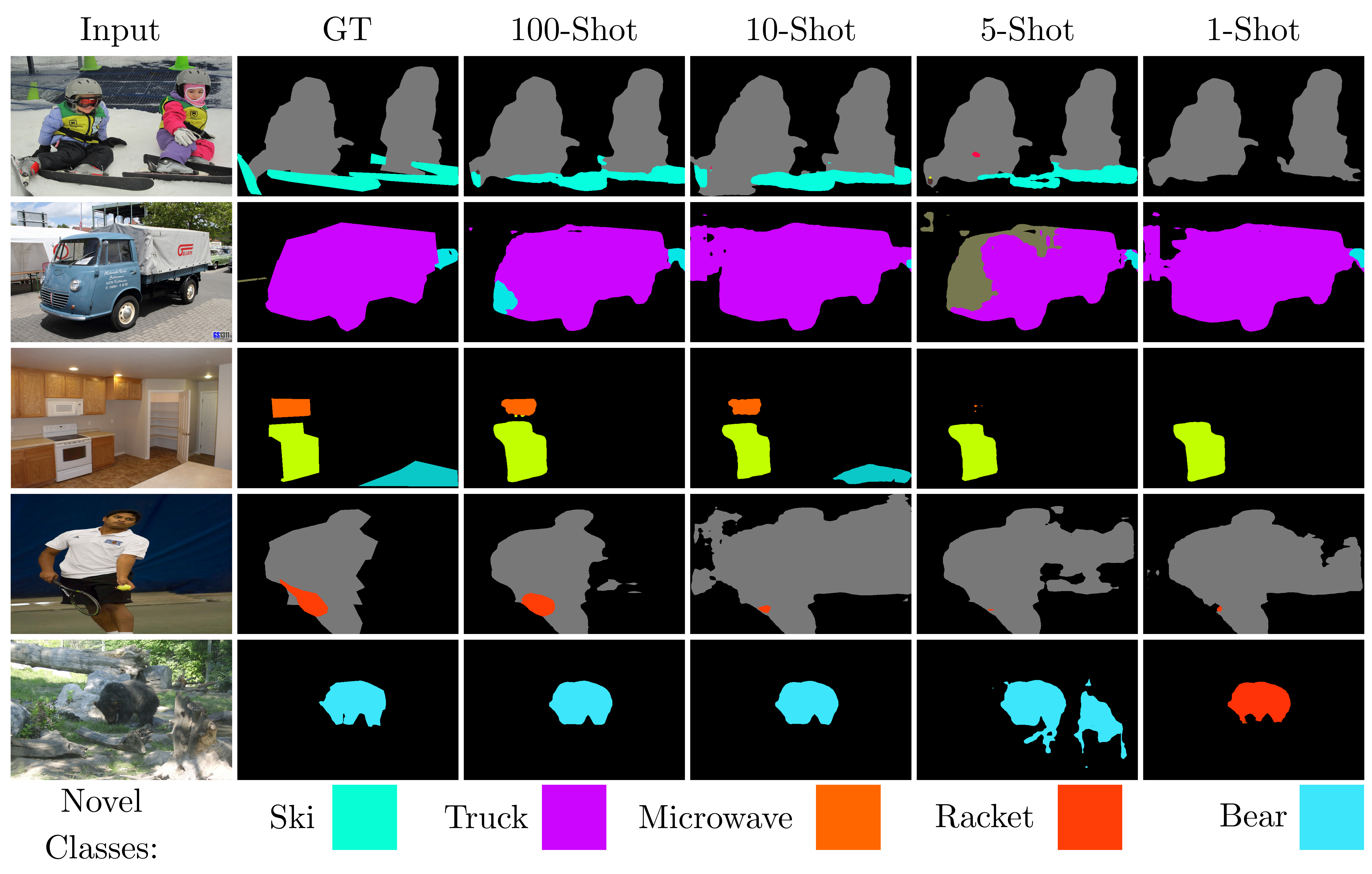}%
\caption{Results for all shots from \emph{Ours-Vanilla} on COCO-$20^i$. As the number of shots increases, the novel class segmentation quality and localization improves.}%
\label{fig: cooc_comp_all}
\end{figure*}

First, we show results from our baseline approach to demonstrate how our approach scales to increasing amounts of data. Specifically, on COCO-$20^i$, we show results from our baseline approach (i.e., \emph{Ours-Vanilla}) for when 1, 5, 10, and 100 shots are available during training. Results are shown in Figure \ref{fig: cooc_comp_all}. We observe better novel class segmentation and identification as the number of shots increase. For example, in the fourth row, the tennis racket is more appropriately segmented from the person when observing 100 shots compared to when only 10 shots are available during training. This reinforces our quantitative findings that our approach continues to learn as the number of shots increases (i.e., we continue to see significant improvements in performance and quality).

\hfill%

\begin{figure*}[!h]
\centering
\includegraphics[width=\textwidth]{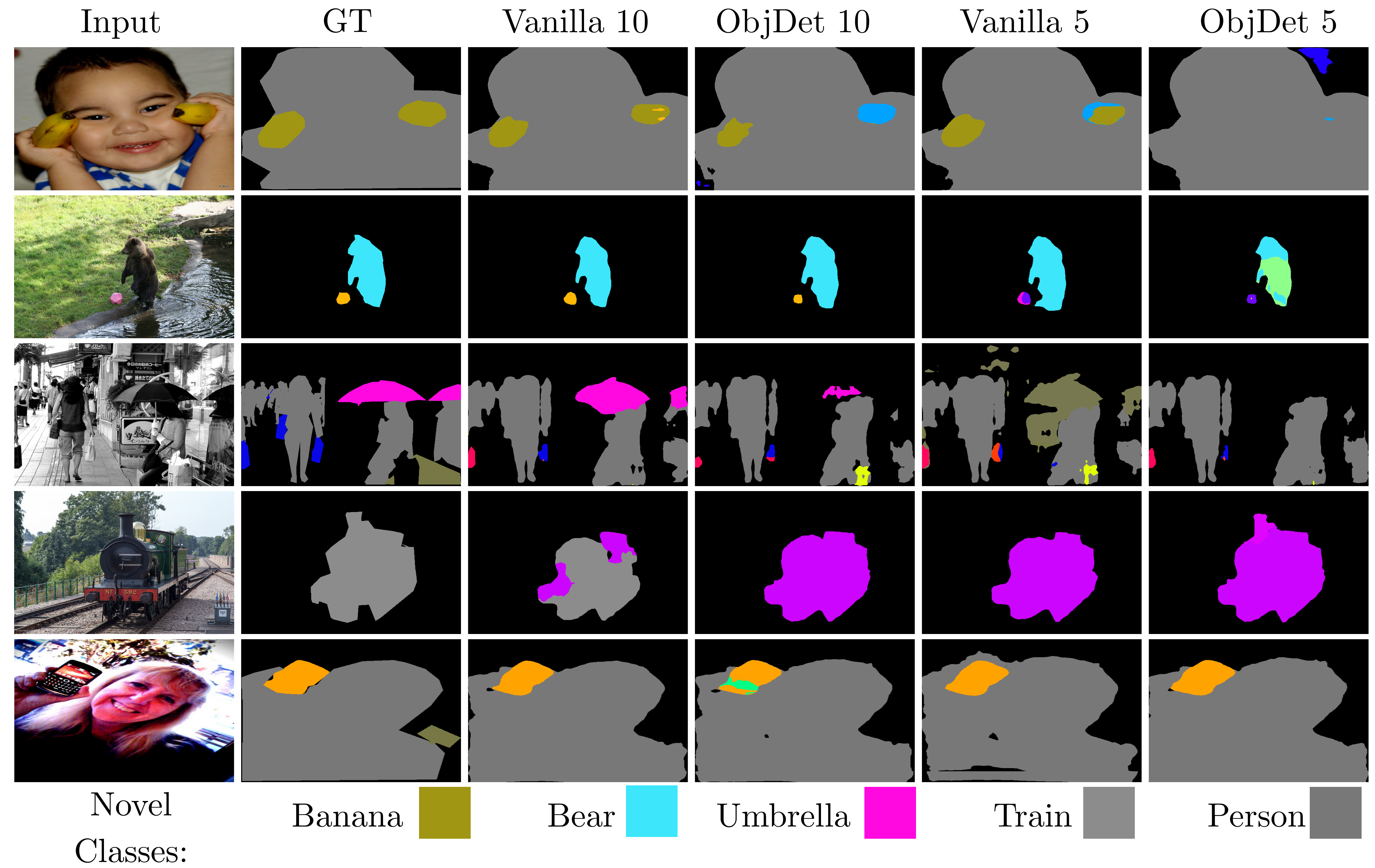}%
\caption{Results for 5 and 10 shots from \emph{Ours-Vanilla} and \emph{Ours-ObjDetFT} on COCO-$20^i$. We continue to observe slightly better novel class identification and segmentation boundaries when fine-tuning more layers (i.e., \emph{Ours-Vanilla}). Note that the novel categories the bottom of the figure correspond to each row.}%
\label{fig: cooc_comp_510}
\end{figure*}

\clearpage
We also show results from different fine-tuning approaches. On COCO-$20^i$, in the 5 and 10-shot settings, we fine-tune: (i) the last layer and (ii) all layers after the backbone. Results are shown in Figure \ref{fig: cooc_comp_510}. These results highlight how fine-tuning more layers results in better segmentation quality for novel classes (e.g., the umbrella in the third row). This finding contradicts the finding of prior work~\cite{wang2020frustratingly} for few-shot object detection that fine-tuning only the last layer is best while also exemplifying how the additional representational power from fine-tuning more layers aids in appropriately identifying novel classes for semantic segmentation.

\hfill%

\begin{figure*}[!h]
\centering
\includegraphics[width=\textwidth]{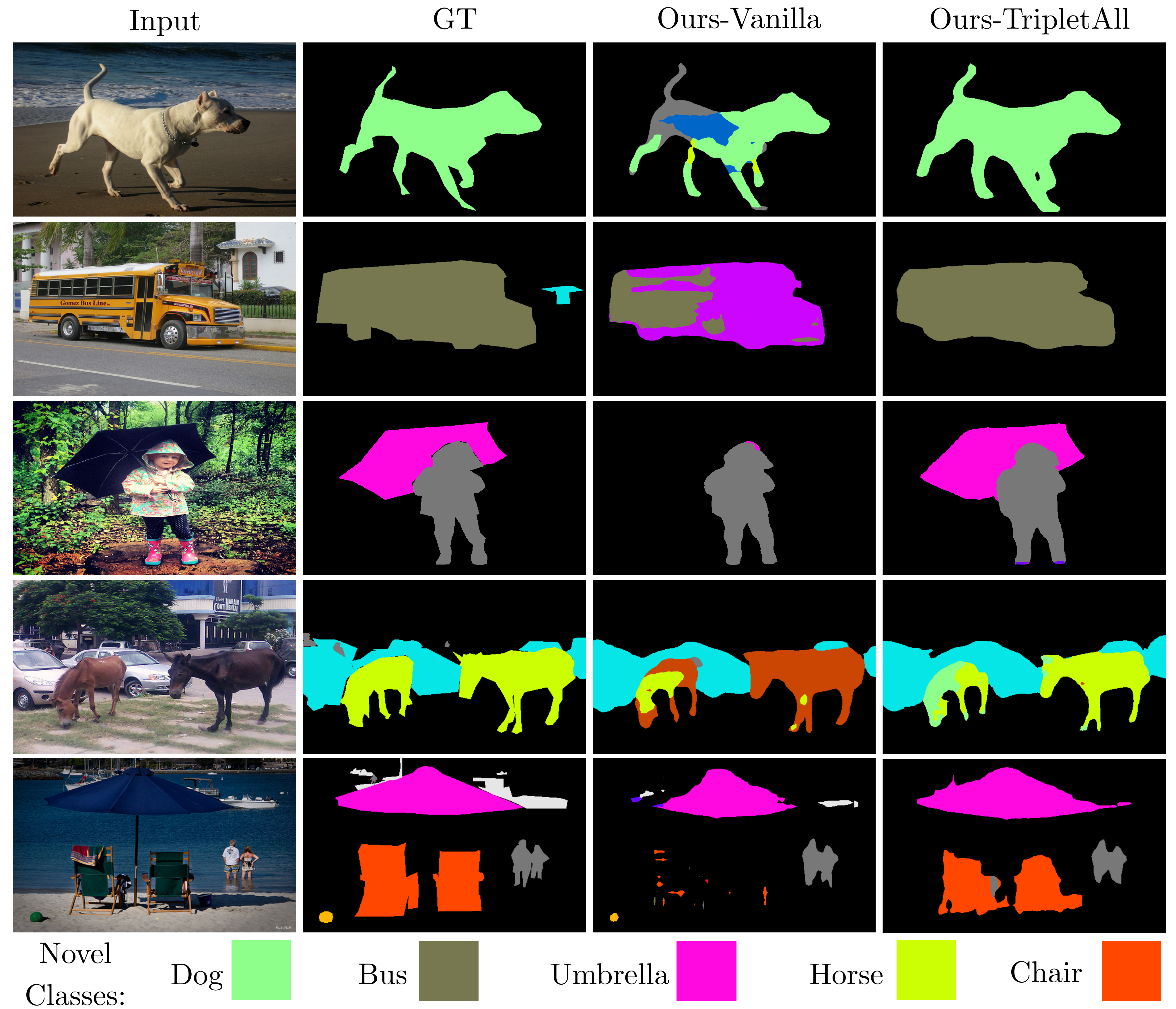}%
\caption{Comparison of \emph{Ours-Vanilla} and \emph{Ours-TripletAll} for COCO-$20^i$. Overall, the addition of triplet loss helps improve the novel class segmentations for both single object and multi-object scenarios.}%
\label{fig: coco_res_ours}%
\end{figure*}

\clearpage
We next exemplify that the inclusion of triplet loss leads to better novel class segmentations. We show results for our baseline approach (i.e., \emph{Ours-Vanilla}) and our augmented triplet loss approach (i.e., \emph{Ours-TripletAll}) approaches for both single and multi-object scenarios on COCO-$20^i$. Results are shown in Figure \ref{fig: coco_res_ours}. These results reinforce our observation from Figure 7 in the main paper that novel classes are segmented better when triplet loss is added to training, with improved outcomes observed both in the single and multi-object scenarios. For example, in the third row, the umbrella is distinguished from the background only by the approach that uses triplet loss. For the same example without triplet loss, the umbrella is barely detected as a present class.

\hfill%

\begin{figure*}[!h]
\centering
\includegraphics[width=\textwidth]{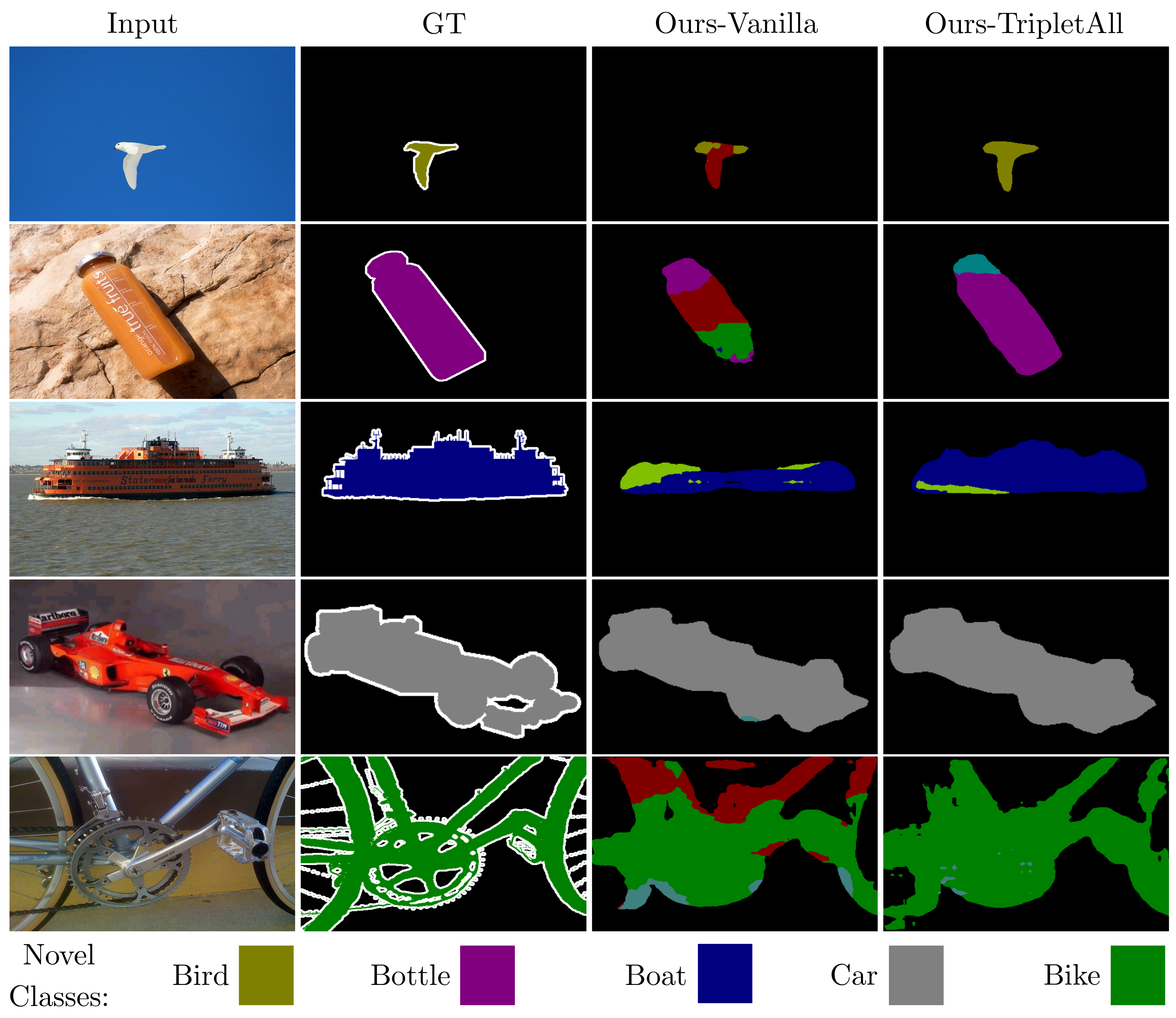}%
\caption{Additional comparisons of \emph{Ours-Vanilla} and \emph{Ours-TripletAll}. Overall, the addition of triplet loss helps improve the novel class segmentations for both single object and multi-object scenarios.}%
\label{fig: res_ours}%
\end{figure*}

\clearpage
Finally, we show additional examples on PASCAL-$5^i$. We compare our baseline approach (i.e., \emph{Ours-Vanilla}) to our baseline approach augmented with triplet loss (i.e., \emph{Ours-TripletAll}) in the 10-shot setting. Results are shown in Figure \ref{fig: res_ours}. With the addition of triplet loss, we observe that novel classes are more appropriately segmented from other classes (i.e., not misclassified) and the overall segmentation quality is improved. The last row highlights this observation as the bike visually appears to only contain bike predictions (i.e., correct predictions) when using triplet loss. In contrast, when triplet is absent (i.e., \emph{Ours-Vanilla}), the last row contains non-bike predictions. This provides further evidence that including explicit similarities between classes (i.e., triplet loss) leads to better class discrimination and segmentation quality compared to only using one supervised loss during fine-tuning (i.e., cross entropy).

\hfill%
 
\clearpage
\subsection{Impact of Novel to Base Class Ratio}
We next examine how the performance of our approach is affected when changing the ratio of novel classes to base classes. We ran experiments on PASCAL-$5^i$ using our baseline approach augmented with triplet loss (i.e., \emph{Ours-TripletAll}) by decreasing the novel class amount by 2 while increasing base classes by 2 (i.e., \emph{Ours-TripletAllLess}) to retain the same number of classes. We also test our approach by increasing novel classes by 2, and decreasing base classes by 2 (i.e., \emph{Ours-TripletAllMore}). We compare these scenarios to our the baseline scenario when a quarter of the classes are novel. 

Results are shown in Table \ref{table: change}. For shots 1, 5, and 10 in the setting of decreased novel classes, total mIoU improves by 6.89, 2.97, and 2.67 percentage points. In the increased setting, total \emph{mIoU} drops by 6.94, 3.60, and 3.18 percentage points compared to our baseline approach augmented with triplet loss. These changes stem largely from base classes in both scenarios, suggesting that our ability to retain base knowledge while introducing novel classes is affected by the ratio of base to novel classes presented during training while novel categories are relatively unaffected.

\begin{table}[!h]
  \begin{center}
  \caption{Comparison between our baseline approach augmented with triplet loss (i.e., \emph{Ours-TripletAll}) and our approach when we decrease the amount of novel classes by 2 (i.e., \emph{Ours-TripletAllLess}), as well as when we increase the amount of novel classes by 2 (i.e., \emph{Ours-TripletAllMore}). We observe changes in performance stemming largely from base classes.} 
  \begin{threeparttable}
    \resizebox{\textwidth}{!}{\begin{tabular}{c c c c c c c c c c }
    \toprule
       \multirow{3}{*}{\bf\shortstack[c]{ \\\\Method}} & \multicolumn{3}{c}{\bf 1-Shot} & \multicolumn{3}{c}{\bf 5-Shot} & \multicolumn{3}{c}{\bf 10-Shot} \\
     \cmidrule(lr){2-4}
     \cmidrule(lr){5-7}
     \cmidrule{8-10}
        & \bf Base & \bf Novel & \bf Total &  \bf Base & \bf Novel & \bf Total & \bf Base & \bf Novel & \bf Total   \\
        \midrule
     \emph{Ours-TripletAll} & 66.41 & 19.71 & 55.31 & 
    71.31 & \bf 50.46 & 66.35 & 
    72.87 & 57.00 & 69.10  \\ 
    \emph{Ours-TripletAllLess} & \bf 69.14 & \bf 20.57 & \bf 62.20 &
        \bf 72.87 & 48.03 & \bf 69.32 &
        \bf 74.03 & \bf 57.98 & \bf 71.74\\
    \emph{Ours-TripletAllMore} & 62.39 & 20.34 & 48.37 &
        68.94 &  50.33 & 62.75 &
        70.99 & 56.03 & 65.92\\
    \bottomrule\\
    \end{tabular}}
    \end{threeparttable}
\label{table: change}
      \end{center}
\end{table}

\subsection{Impact of Novel-Only Evaluation}
We conduct experiments to demonstrate the performance of our approach in the traditional few-shot setting. We compare how our approach performs in the $n$-way, $k$-shot scenario where only novel classes must be identified (i.e., \emph{Ours-Traditional}) to our baseline approach (i.e., \emph{Ours-Vanilla}) where we must identify both base and novel classes during training.  

Results are shown in Table \ref{table: novel}.  We observe a decrease in performance (i.e., novel \emph{mIoU}) when only evaluating on novel classes, compared to our baseline approach.  We do not find this surprising, since our method is designed to retain knowledge of base categories and so we would expect it to perform worse than methods designed to only perform well on novel classes.

\begin{table}[!t]
  \begin{center}
  \caption{Comparison between our approach in the generalized few-shot semantic segmentation setting (i.e., \emph{Ours-Vanilla}) vs. our approach in the traditional few-shot setting (i.e., \emph{Ours-Traditional}) on PASCAL-$5^i$. We observe worse results when only evaluating on novel categories as our method is designed to retain base category knowledge.} 
  \begin{threeparttable}
    \begin{tabular}{c c c c  }
    \toprule
       \multirow{3}{*}{\bf\shortstack[c]{ \\\\Method}} & \multicolumn{1}{c}{\bf 1-Shot} & \multicolumn{1}{c}{\bf 5-Shot} & \multicolumn{1}{c}{\bf 10-Shot} \\
     \cmidrule(lr){2-2}
     \cmidrule(lr){3-3}
     \cmidrule{4-4}
        & \bf Novel &  \bf Novel & \bf Novel  \\
        \midrule
     \emph{Ours-Vanilla} & \textbf{18.82} & 
   \textbf{46.40} & \textbf{52.55} \\ 
     \emph{Ours-Traditional} & 13.86  & 38.64 &  45.42 \\
    \bottomrule\\
    \end{tabular}
    \end{threeparttable}
\label{table: novel}
      \end{center}
      \vspace{-3em}
\end{table}

\begin{table}[!t]
  \begin{center}
  \caption{Comparison on PASCAL-$5^i$ between two fine-tuning approaches: (i) only the last layer in the network (i.e., \emph{FSDet-Last}) and (ii) fine-tuning all layers after the backbone (i.e., \emph{FSDet-Backbone}) for few-shot object detection. We observe worse performance when fine-tuning more layers. Performance is measured as Average Precision 50 for base (i.e., bAP50), novel (i.e., nAP50) and total (i.e., AP50).} 
  \begin{threeparttable}
    \resizebox{\textwidth}{!}{\begin{tabular}{c c c c c c c c c c }
    \toprule
       \multirow{3}{*}{\bf\shortstack[c]{ \\\\Method}} & \multicolumn{3}{c}{\bf 1-Shot} & \multicolumn{3}{c}{\bf 5-Shot} & \multicolumn{3}{c}{\bf 10-Shot} \\
     \cmidrule(lr){2-4}
     \cmidrule(lr){5-7}
     \cmidrule{8-10}
        & \bf bAP50 & \bf nAP50 & \bf AP50 &  \bf bAP50 & \bf nAP50 & \bf AP50 & \bf n5AP0 & \bf nAP50 & \bf AP50   \\
        \midrule
     \emph{FSDet-Last}~\cite{wang2020frustratingly} & \bf 88.36 & \bf 19.85 & \bf 71.24 &
    \bf 88.17 & \bf 43.92 & \bf 77.11 &
    \bf 87.70 & \bf 52.88 & \bf 78.99\\
     \emph{FSDet-Backbone}~\cite{wang2020frustratingly} & 81.49 & 18.79 & 65.82 &
     77.60 & 43.28 & 69.02 &
     78.41 & 51.21 & 71.61\\
    \bottomrule\\
    \end{tabular}}
    \end{threeparttable}
\label{table: fsdet_comp}
      \end{center}
\end{table}

\section{Few-Shot Object Detection Fine-Tuning Approaches}
Now we validate that few-shot object detection~\cite{wang2020frustratingly} experiences different performance when fine-tuning different layers. We ran experiments on PASCAL-$5^i$ to compare: (i) fine-tuning only the last layer, and (ii) fine-tuning all layers after the backbone. In order to observe stable results, we average our results over 3 seeds for 1, 5, and 10 shots. Results are shown in Table \ref{table: fsdet_comp}. We observe a performance drop when fine-tuning more layers, contrasting what we observe in generalized few-shot semantic segmentation (i.e., fine-tuning more layers led to better performance). Given that object detection produces less dense outputs than semantic segmentation (i.e., bounding box vs. per-pixel classifications), there may be sufficient representational power in the final layer such that fine-tuning more parameters leads to under-fitting.